\documentclass[journal,twoside,web]{ieeecolor} 
\usepackage{tmi}

\usepackage{marvosym}
\usepackage[rgb,x11names]{xcolor}
\usepackage{graphicx}
\usepackage{amsfonts,amsmath,amssymb}
\usepackage[tight]{subfigure}
\usepackage{multirow}
\usepackage{subfigure}
\usepackage{pgfplots}
\usepackage{tikz-3dplot}
\usepackage{cite}

\usepackage[colorlinks=true,linkcolor=blue,citecolor=blue]{hyperref}

\usepackage{breqn}
\usepackage{booktabs}
\usepackage{todonotes}
\usepackage{soul}

\usepackage{paralist}
\usepackage{enumerate}
\newcommand{\rev}[1]{\textcolor{black}{#1}}

\usepackage{amsmath}
\usepackage[ruled,vlined,boxed,linesnumbered]{algorithm2e}
\SetKwInOut{KwIn}{Server Input}
\SetKwInOut{KwOut}{Server Output}
\SetKwData{Data}{Hospital $k$'s Input}
\SetKwData{Result}{Hospital $k$'s Output}

\usepackage{algpseudocode}

\pgfdeclarelayer{background}
\pgfdeclarelayer{foreground}
\pgfsetlayers{background,main,foreground}

\pgfplotsset{compat=1.3}

\usepackage{pifont}%
\newcommand{\mybox}{\fbox}
\tikzstyle{fancylabel} = [text=yellow, inner sep=0pt, minimum size=15pt, align=left]
\tikzstyle{mylabel} = [text=white, ultra thick, inner sep=1pt, minimum size=15pt, yshift=-4pt, xshift=6pt]

\def\ourmodel{MFed-CBT}
\def\ourmodel{\rev{MetaFedCBT}}
\def\cbt{DGN}
\label{packages}
\usepackage{environ}
\renewcommand{\mybox}{}
\newcommand{\mytodo}[1]{\todo{TODO}}

\def\BibTeX{{\rm B\kern-.05em{\sc i\kern-.025em b}\kern-.08em
		T\kern-.1667em\lower.7ex\hbox{E}\kern-.125emX}}
\begin{document}

\title{Metadata-Driven Federated Learning of Connectional Brain Templates in Non-IID Multi-Domain Scenarios}



\author{
	Geng Chen\thanks{\Funding}\thanks{\NWPU}, 
	Qingyue Wang, and Islem Rekik, \IEEEmembership{Member, IEEE}\thanks{\ICL}
	\thanks{\CorrAuthor}
	}

\def\Funding{This work was funded by generous grants from the National Natural Science Foundation of China (grant no. 62201465) to G. Chen, the European H2020 Marie Sklodowska-Curie action (grant no. 101003403, \url{http://basiralab.com/normnets/}) to I. Rekik, and the Scientific and Technological Research Council of Turkey to I. Rekik under the TUBITAK 2232 Fellowship for Outstanding Researchers (no. 118C288, \url{http://basira-lab.com/ reprime/}).
However, all scientific contributions made in this project are owned and approved solely by the authors.}

\def\NWPU{G.~Chen and Q.~Wang are with National Engineering Laboratory for Integrated Aero-Space-Ground-Ocean Big Data Application Technology, School of Computer Science and Engineering, Northwestern Polytechnical University, China.}

\def\ICL{I.~Rekik is with BASIRA Lab \url{https://basira-lab.com/}, Imperial-X and Department of Computing, Imperial College London, United Kingdom.}

\def\CorrAuthor{Corresponding Author: I.~Rekik (Email: i.rekik@imperial.ac.uk).}




\maketitle

\begin{abstract}
A connectional brain template (CBT) is a holistic representation of a population of multi-view brain connectivity graphs, encoding shared patterns and normalizing typical variations across individuals. The federation of CBT learning allows for an inclusive estimation of representative center of multi-domain brain connectivity datasets in a fully data-preserving manner. However, existing methods overlook the non-independent and identically distributed (non-IDD) issue stemming from multi-domain brain connectivity heterogeneity, in which data domains are drawn from different hospitals and imaging modalities. This non-IDD issue degrades the centrality of locally learned CBT from multiple decentralized datasets (i.e., domains/hospitals), eventually leading to a limited representation ability.
To overcome this limitation, we unprecedentedly propose a {metadata-driven} federated learning framework, called \ourmodel{}, for cross-domain CBT learning. Given the data drawn from a specific domain (i.e., hospital), our model aims to learn a metadata \emph{in a fully supervised manner} by introducing a local client-based regressor network. The generated meta-data is forced to meet the statistical attributes (e.g., mean) of \emph{other} domains, while preserving their privacy. Our supervised meta-data generation approach boosts the unsupervised learning of a more centered and representative and holistic CBT of a particular brain state across diverse domains. As the federated learning progresses over multiple rounds, the learned metadata and associated generated connectivities are are continuously updated to better approximate the target domain information.
\ourmodel{} overcomes the non-IID issue of existing methods by generating informative brain connectivities for privacy-preserving holistic CBT learning with the guidance using metadata.
Extensive experiments on multi-view \rev{morphological} brain networks of normal and patient subjects demonstrate that our \ourmodel{} is a superior federated CBT learning model and significantly advances the state-of-the-art performance.
\end{abstract}


\section{Introduction}
\begin{figure}[t]
	\centering
	\mybox{\includegraphics[trim={0cm 8.3cm 17.5cm 0cm},clip,width=0.95\linewidth,page=1]{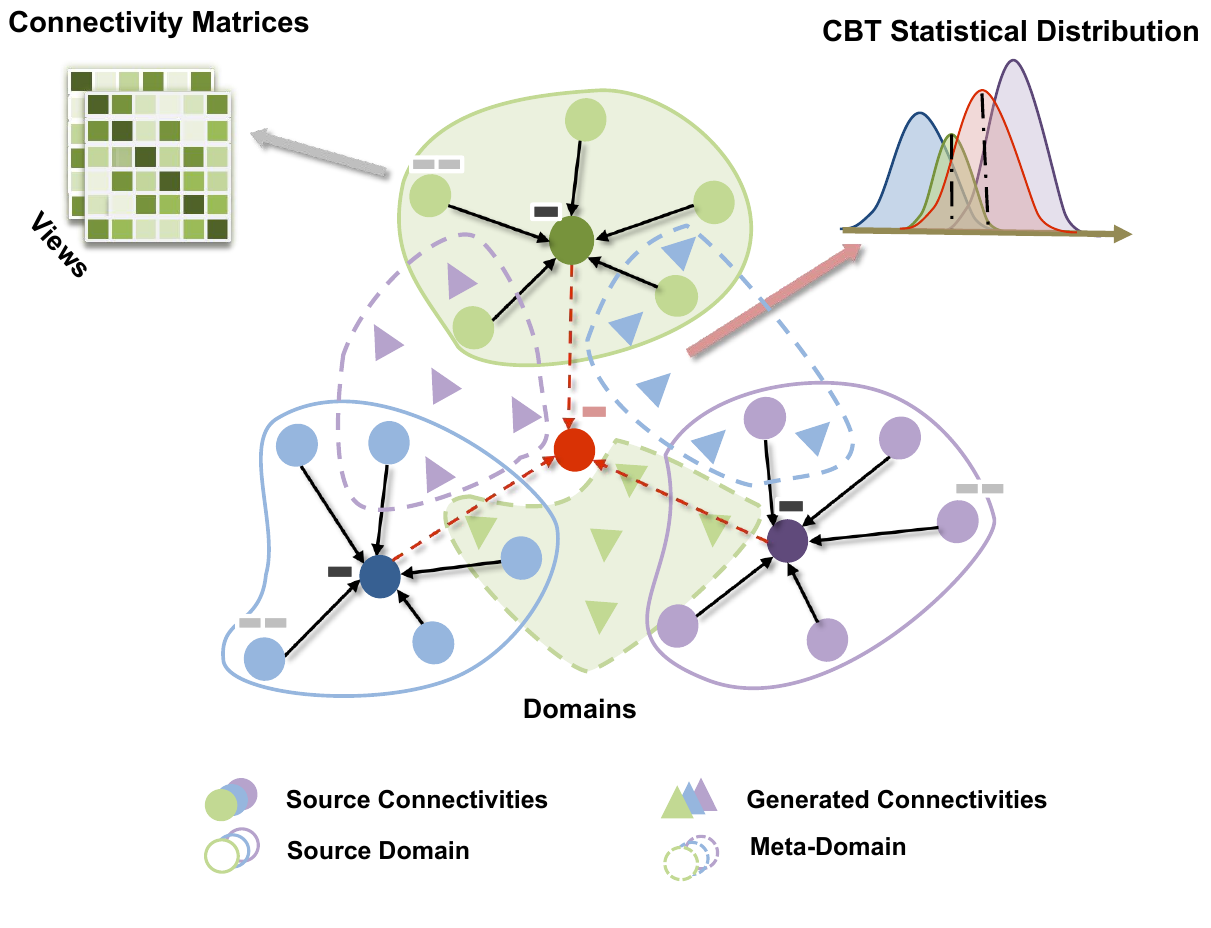}}
 
	\vspace{-3pt}
\caption{Illustration of heterogeneous multi-view brain connectivities drawn from different hospitals (i.e., domains) for federated learning of holistic CBT across multiple domains, each marked by a corresponding color.
To address the non-IID issue associated with the source domains (circled by the solid lines), we propose a data-generation federated learning scheme, where the meta-domains marked by the dashed circled lines are generated to shift the locally learned CBTs towards a more centered template across all federation domains.
}\label{fig:introduction}
\end{figure}
\IEEEPARstart{A}{} connectional brain template (CBT) is the representative center of a population of multi-view brain connectivities \cite{2019Estimation,chaari2023comparative}.
It plays a vital role in understanding the patterns of brain connectivity across different subjects and imaging modalities in health and disease \cite{chaari2023comparative} since it can provide a connectional representation that is well-centered, discriminative, and topologically sound of a particular brain state (e.g., healthy or disordered). A straightforward solution to constructing a CBT is linear averaging, which is, however, unable to capture the nonlinear complex variability across subjects and views (or modalities).
Graph representation depicts complex multi-subject multi-view brain connectivities. Specifically, by representing brain connectivities as graphs, one can leverage powerful graph-learning techniques to estimate CBTs from populations of brain graphs.
For instance, multigraph integration methods \cite{2019Clustering,2020Estimation,2020Deep,2021MGN} are employed to fuse the multi-view connectivity data into a single-view representation.

Despite its recent advances, CBT learning is still a challenging task, especially when dealing with a large scale of multi-domain (i.e., multi-center) data collected from different sources. This is mainly due to the privacy issue associated with multi-domain medical data\cite{2020FedNER}.
To address this challenge, recent efforts have been dedicated to {tackle} multi-domain CBT learning using powerful federated learning (FL) techniques, which learn from multiple {distributed} hospitals without accessing the data stored in these centers. A typical FL technique is FedAvg\cite{FedAvg}, which trains distributed models independently on each data owner (a.k.a. client). The trained models are sent to a central server, which aggregates the updates for training the global model. This ``server-client'' paradigm has significant advantages over traditional centralized learning, especially in settings where data privacy and security are of utmost importance, such as healthcare.

The first attempt at federating the learning of CBT is Fed-CBT \cite{2021A}, which combines FedAvg with a multi-view graph integration method \cite{2020Deep}.
Fed-CBT employs a baseline FL algorithm to learn a CBT by federating the deep graph normalizer (DGN) \cite{2020Deep} architecture across participating hospitals trained on their local multiv-view datasets. As such, the learned CBT captures shared connectivity patterns across multi-source data in a holistic way with higher communication efficiency. Although making significant progress in cross-domain CBT learning, Fed-CBT suffers from a number of challenges, the major one lies in the non-independent and identically distributed (non-IID) data distribution \cite{li2020federated}, as shown in Fig.~\ref{fig:introduction}.
The non-IID data have different statistical properties across different domains (i.e., hospitals), such as different data distributions, feature spaces, or label spaces. In practice, the brain connectivity data collected at different hospitals are non-IID due to the inherent heterogeneity in acquisition protocols and population characteristics, as well as imaging modality (e.g., T1-w MRI, diffusion MRI). Each hospital brings a unique domain-specific representative to the cohort of brain connectivities, capturing specific aspects of the complex brain graph architecture. The non-IID nature of the data leads to challenges in generalization, where the locally learned CBTs may lack global representativeness or even diverge during the training process. This hinders the ability to capture the true variable and shared patterns of brain connectivity across diverse populations and undermines the overall effectiveness of the CBT learning model. 

In this paper, we propose a novel \emph{metadata-driven domain-generation FL} framework, called \ourmodel{}, to learn a CBT from non-IID brain connectivities, which are drawn from disparate domains (i.e., hospitals). In a multi-hospital scenario, a hospital-specific learned CBT is fully agnostic to the data of other hospitals, {which limits its centerdness and representativeness to only its local data distribution.} 
Our goal is to learn a universal CBT that approximates a global center across various brain connectivity types (e.g., functional and morphological) or domains without data sharing. 
To alleviate the cross-domain heterogeneity problem, we improve the generalization of the local model (trained at each hospital) by boosting the data diversity of each hospital using a custom strategy of supervised meta-domain data generation.
This lies in the assumption that \emph{if a hospital learns how to predict the distribution of other centers, it can contribute better to generating a more representative template of multi-view brain graphs}. 
Particularly, we analyze this distributional information from round-by-round exchanged weights between global (at the server level) and local models (at the hospital/client level).
In other words, the aggregated FL global model carries multiple domain information that can be directly accessed by clients during the communication of model weight parameters without data sharing.
Specifically, we introduce the concept of metadata to represent such an unseen latent space, namely \emph{meta-domain}, by defining the metadata as the key statistical attributes of a meta-domain (e.g., mean and standard deviation values). 
The brain connectivity graphs follow a normal distribution, allowing statistical information to guide the generation of a new sample (i.e., brain graph). 
Specifically, we \emph{supervisedly} predict the metadata using a distribution regressor with the network residuals (i.e., the difference between global and local model weights). Our designed regressor maps the network residuals (i.e., the difference between global and local model weights) onto the metadata (i.e., statistics of data at the meta-domain). The resulting metadata is then used to generate high-quality multi-view connectivity data (i.e., multigraphs) by the trained generator. The connectivities are locally generated and then utilized to improve the quality of the learned CBT.
This two-step ``FL + data generation'' approach is iteratively conducted to learn a powerful global FL model for improving federated CBT learning. The details of our regressor training strategy are detailed in the Method Section.

In summary, our contributions are four-fold:
\begin{enumerate}
	\item Our \ourmodel{} is the first FL framework for unsupervised integration of multi-graph brain data non-independent and identically distributed across disparate domains, for learning a more centered and representative CBT of a particular brain state.
	\item We propose a novel supervised distribution regressor based on a round-update residual network to predict hospital-wise metadata that approximates the distributions of unseen domains. This allows for moving each locally learned CBT towards a more global multi-domain center, thereby generating a universal CBT through federation.
	\item {We generate informative brain connectivities in each source domain with the supervision of the predicted metadata, overcoming the non-IID challenge.}
        \item We perform extensive experiments on the multi-view connectomes from the Autism Brain Imaging Data Exchange (ABIDE-I) dataset. The results show that our \ourmodel{} outperforms state-of-the-art models remarkably.
\end{enumerate}


\section{Related Work}
\label{section2}
\subsection{CBT learning}
The goal of CBT learning is to fuse several brain connection networks that were obtained from various neuroimaging modalities or brain perspectives (e.g., structural and functional views) into a single integrated graph. In a more abstract way, note that this is an unsupervised learning task where we set out to learn a latent center of a highly heterogeneous multimodal population of multigraphs. It can also be expressed as a multimodal/multiview graph integration or fusion task.
The CBT aligns brain networks based on connectivity patterns, which can be mapped using multigraph integration approaches.
For instance, Dhifallah et al. \cite{2019Clustering} proposed a clustering-based unsupervised multi-view fusion network, called SCA for CBT learning by fusing different views of each subject into one view and then clustering these views based on the similarity of their mapped connectivity patterns.
Further, Dhifallah et al. \cite{2020Estimation} built a higher-order graph of ROIs across views that captures the variability in connectivity patterns across individuals, then selects the most central connection tensor with a graph-based feature selection method.
NetNorm \cite{2020Estimation} is an advanced unsupervised CBT learning method, but might cause the accumulation of errors in the estimation pipeline due to independent learning steps.
Deep graph normalizer (DGN) \cite{2020Deep} presents the first graph neural network (GNN) for multigraph fusion. It addresses both cross-view and cross-individual variability and complexity of a given population. Specifically, each layer of GNN integrates views and learns deeper embeddings for each node, generating the most well-centered CBT based on a random subject normalized loss (SNL). DGN presents the state-of-the-art method in multi-view data integration according to a recent comparative survey \cite{chaari2023comparative}. MGN-Net \cite{2021MGN} further improves DGN by introducing a new loss function, which preserves the topology of the learned CBT. 

Although effective, such methods were not originally designed to learn a CBT from graph data distributed at different centers. To overcome this limitation, the effort has been recently dedicated to developing multi-center CBT learning using powerful FL algorithms, which utilize knowledge from several remote centers without accessing the data kept in these centers.
A pioneering approach is Fed-CBT \cite{2021A}, which is an FL framework based on a multi-view graph connectivity mapping approach that attempts to build a globally representative cross-hospital CBT in a completely data-preserving way. However, TW-FedCBT \cite{2021A} is designed to improve the FL aggregation by a temporally weighted strategy, failing to handle the severe heterogeneous case where the data across hospitals is non-IIDs as it is limited to learning from hospitals with similar domain distributions.
Precisely, the template generated by Fed-CBT may not fully capture the inter-individual brain connection variability in a federated setting.

\subsection{Federated Learning in Medical Imaging}
FL is a distributed machine learning approach that enables multiple parties to train a shared machine learning model without sharing their raw data \cite{FedAvg}.
To achieve this, the parties train their local models on their respective datasets and share only the model updates with a central server.
The central server aggregates these updates to update the shared model, which can then be deployed on local devices for local inference.

In medical applications, a neural network is trained using a medical imaging dataset collected from multiple hospitals in an FL scenario \cite{10077569}. 
There are several advantages for FL in medical imaging applications \cite{sarma2021federated}. First, FL enables the collaborative training of machine learning models across different institutions without sharing sensitive patient data. This is particularly important for medical imaging, where the privacy and security of patient data are of utmost importance. Second, it allows for the efficient use of distributed resources, enabling the training of large-scale machine-learning models on distributed data. This can improve the accuracy and robustness of machine-learning models for medical imaging applications. Third, FL eliminates the need for large-scale data transfers, which can save bandwidth and reduce latency. 

Thanks to these advantages, FL has been widely applied in medical imaging, such as disease detection \cite{Adnan2021FederatedLA,pati2022federated,nguyen2021federated}, medical image segmentation \cite{nalawade2021federated,liu2021feddg,yang2021federated}, medical Image classification \cite{yan2020variation,liu2021federated,luo2022fedsld}, and brain network analysis \cite{2021A,gurler2022federated}. 
Interested readers can refer to \cite{sohan2023systematic} for a comprehensive review of FL in medical imaging.
{However, FL also poses several challenges that need to be addressed to ensure the quality, privacy, and security of the machine learning models.
In particular, FL faces a significant challenge in non-IDD scenarios, where the FL performance degrades. Multi-domain learning (MDL) within the context of FL is an emerging research area that aims to address the challenges of FL in scenarios with heterogeneous and non-IID data distribution across different domains.
{Li et al. \cite{li2021fedbn} proposed to keep batch normalization layers locally during model aggregation.}
Sun et al. \cite{sun2021partialfed} offered a new solution to address performance degradation in cross-domain FL by loading only a subset of the global model's parameters.
Wicaksana et al. \cite{wicaksana2022customized} introduced a customized FL, where clients iteratively train individual models based on global feature extraction layers.
This method outperforms methods that use the federated model for feature alignment and guide private model training, enabling clients to enhance personalized models.}

Existing multi-domain FL strategies have only been validated in supervised or semi-supervised tasks, such as image classification.
For our \emph{unsupervised} CBT learning task, a specialized design is required. 
To address this challenge, we propose a metadata-driven FL framework for more accurate CBT learning. For each hospital, our \ourmodel{} generates brain connectivity data to learn a meta-domain, which provides complementary information for the unseen hospitals and significantly alleviates the non-IDD issue.

\section{Method}
\label{section3}
\subsection{Metadata-Driven Federated CBT learning}

\begin{figure*}[t]
\centering	\mybox{{\includegraphics[trim={0cm 6.0cm 11.8cm 0cm},clip,width=0.99\linewidth,page=3]{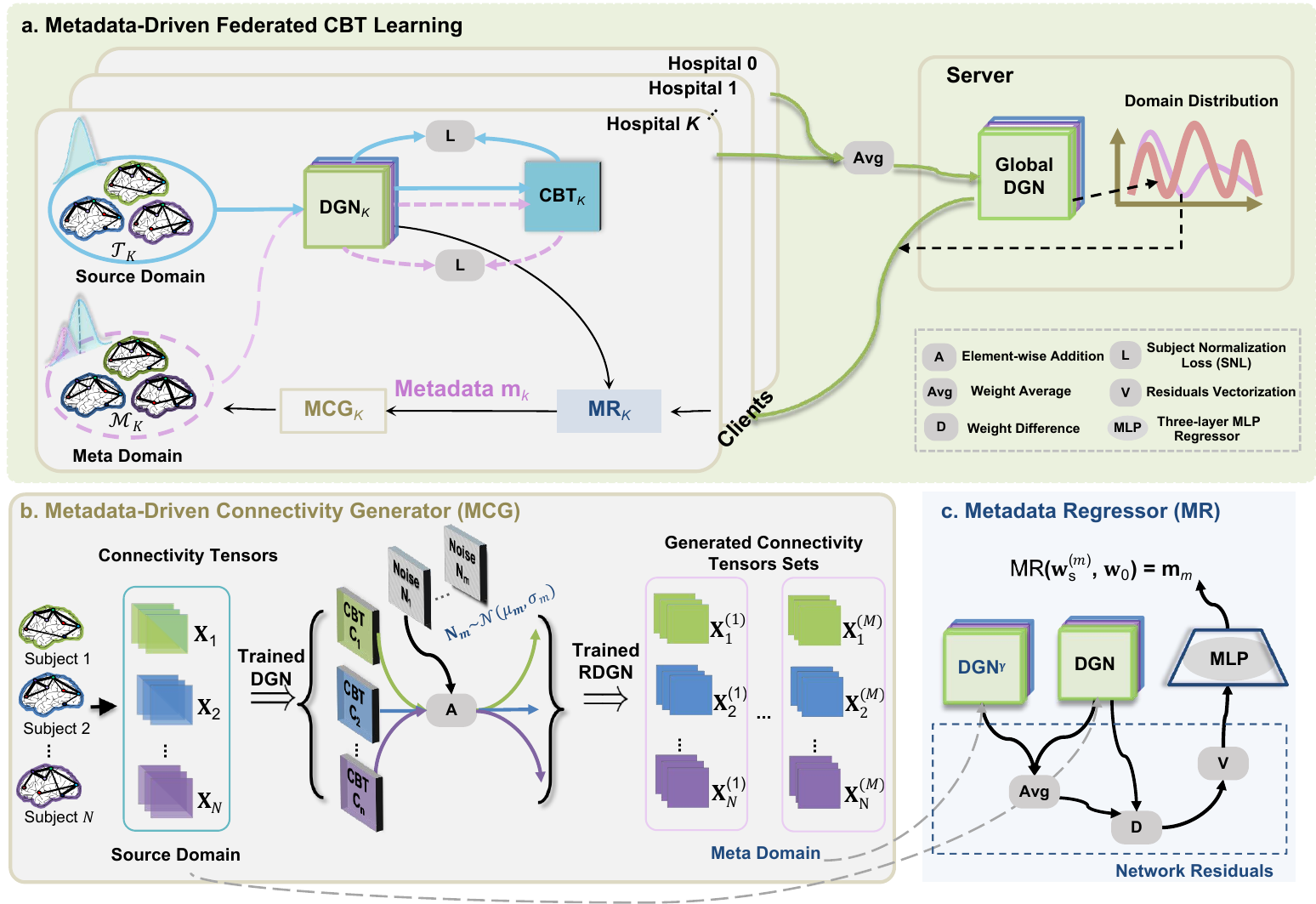}}}
\vspace{-3pt}
\caption{{An overview of \ourmodel.} {(a) indicates how metadata is incorporated into FL using the metadata-driven connectivity generators $\texttt{MCG}_k$ and metadata regressors $\texttt{MR}_k$ for improved performance.
For each hospital, we predict its metadata $\textbf{m}$ with a metadata regressor (c) from corresponding network residual weights between the local hospital and the server during the FL.
{The statistical property is then employed to determine a symmetric Gaussian noise matrix $\mathbf{N} \sim \mathcal{N}({\mu},{\sigma})$ to increase disturbance on the original multi-view dataset $\mathcal{T}$.}
This facilitates the generation of a meta-domain $\mathcal{M}$ through a metadata-driven connectivity generator (b) for each hospital.
Thanks to the high-quality meta-domain generated in a supervised manner, our \ourmodel{} effectively improves the FL of CBTs.}}
\label{fig:overview}
\end{figure*}

\begin{algorithm}[t]
	\caption{{Federated learning of CBT with Brain Connectivities from $K$ Hospitals.}}\label{alg:fedcbt}
        \textbf{Server Input}: The DGN weights $\mathbf{\textbf{w}}_k$ of hospital $k$
        
        \textbf{Hospital $k$'s Input}: Initial global DGN model weights\ $\mathbf{\textbf{w}}$, local training data $\mathcal{T}_k$, brain graphs for hospital $k$, number of rounds $T$, number of rounds per models averaging $E$,number of samples at all $K$ hospitals $N$, $N_k$ is the number of samples at $k$th hospital
        
        \SetKwFunction{MyFunction}{Hospitals} 
        \SetKwProg{Fn}{}{:}{\KwRet} 
       \Fn{\MyFunction}{

	\For{$t$ in $0,\cdots,T-1$}{
		\For{$k$ in $1, 2, \cdots, K$}{
             \textbf{Initialize} local model using $\mathbf{\textbf{w}}_k^{(t)}=\mathbf{\textbf{w}}^{(t)}$\;
	   \textbf{Update} network weights using $\mathbf{\textbf{w}}_k^{(t+1)} =\texttt{DGN}(\mathbf{\textbf{w}}_k^{(t)}, \mathcal{T}_k, E)$\;
            \textbf{Communicate} $\mathbf{\textbf{w}}_k^{(t+1)}$ to the server\;
            }
            } }
    \SetKwFunction{MyFunction}{Server} 
        \SetKwProg{Fn}{}{:}{\KwRet} 
     \Fn{\MyFunction}{
    \textbf{Average} model weights using $\mathbf{\textbf{w}}^{(t+1)}= \sum_{k=1}^K\frac{N_k}{N}\mathbf{\textbf{w}}_k^{(t+1)}$\;
    \textbf{Communicate} $\mathbf{\textbf{w}}^{(t+1)}$ to all the hospitals\;}
    \textbf{Server output}: $\mathbf{\textbf{w}}^{(t+1)}$ to hospital $k$

    \textbf{Hospital $k$'s output}: $\mathbf{\textbf{w}}_k^{(t+1)}$  
\end{algorithm} 

\begin{algorithm}[t]
    \caption{{Metadata-Driven Federated learning of CBT with Brain Connectivities from $K$ Hospitals.}}\label{alg:medfed}
    \textbf{Server Input}: The DGN weights $\mathbf{\textbf{w}}_k$ of hospital $k$
    
    \textbf{Hospital $k$'s Input}: Initial global DGN model weights $\mathbf{\textbf{w}}$, constructed CBTs $\{\textbf{C}_k\}_{k=1}^{K}$ with pretrained $\texttt{MCG}_k$, local training data $\mathcal{T}_k$: brain graphs for hospital $k$, number of rounds $T$, number of rounds per models averaging $E$, number of samples at all $K$ hospitals $N$, $N_k$ is the number of samples at $k$th hospital
    
    \SetKwFunction{MyFunction}{Hospitals} 
    \SetKwProg{Fn}{}{:}{\KwRet}

    \Fn{\MyFunction}{
    \textbf{Construct} CBTs randomly perturbed on each raw multigraph $\{\mathbf{\hat{C}}_k\}_{k=1}^{K}$\;
    \textbf{Train} the metadata regressors $\{\texttt{MR}_k(\cdot,\cdot,\cdot)\}_{k=1}^{K}$ with sets of generated brain connectivities through connectivity generator with disturbed CBTs\;
    \For{$t$ in $0,\cdots,T-1$}{
        \For{$k$ in $1, 2, \cdots, K$}{
         \textbf{Initialize} local model $\mathbf{\textbf{w}}_k^{(t)}=\mathbf{\textbf{w}}^{(t)}$, local training data $\mathcal{\hat{T}}_k= \mathcal{T}_k$\;
         \textbf{Update} network weights using $\mathbf{\textbf{w}}_k^{(t+1)} =\texttt{DGN}(\textbf{w}_k^{(t)}, \mathcal{\hat{T}}_k, E)$\;
            \textbf{Communicate} $\mathbf{\textbf{w}}_k^{(t+1)}$ to the server\; 
        \For{$k$ in $1, 2, \cdots, K$}{
		\textbf{Estimate} metadata using $\textbf{m}_k^{(t+1)} =\texttt{MR}_k(\textbf{w}_k^{(t+1)},\textbf{w}^{(t+1)})$\;
		 \textbf{Generate} brain connectivities using $\mathcal{M}_k^{(t+1)} = \texttt{MCG}_k(\textbf{m}_k^{(t+1)}, \textbf{C}_k)$\;
		\textbf{Update} training data using $\mathcal{\hat{T}}_k = \mathcal{T}_k \bigcup \mathcal{M}_k^{(t+1)}$\;
        }
      }
    }
    }
      \SetKwFunction{MyFunction}{Server} 
        \SetKwProg{Fn}{}{:}{\KwRet} 
     \Fn{\MyFunction}{
     \textbf{Average} model weights using $\mathbf{\textbf{w}}^{(t+1)}= \sum_{k=1}^K\frac{N_k}{N}\mathbf{\textbf{w}}_k^{(t+1)}$\;
        \textbf{Communicate} $\mathbf{\textbf{w}}^{(t+1)}$ to all the hospitals\;
     }
       \textbf{Server output}: $\mathbf{\textbf{w}}^{(t+1)}$ to hospital $k$
       
        \textbf{Hospital $k$'s output}: $\mathbf{\textbf{w}}_k^{(t+1)}$, $\mathcal{\hat{T}}_k^{(t+1)} $
\end{algorithm}

An overview of our \ourmodel{} is shown in Fig.~\ref{fig:overview}. As can be observed, \ourmodel{} constructs the brain CBT with distributed \cbt{} \cite{2020Deep} models in an FL manner. A major innovation distinguishing \ourmodel{} from conventional federated CBT learning lies in the design of the metadata, which significantly improves the unsupervised learning of a CBT by addressing the non-IID problem associated with multi-domain brain connectivities from different hospitals.

\subsubsection{Revisiting the Federated CBT learning}
Before introducing the details of \ourmodel{}, we would like to revisit the federated CBT learning \cite{2021A}, which can help understand the details of \ourmodel{}. First of all, we propose our \ourmodel{} to solve the problem of holistic multi-view connectivity mapping across multiple hospitals based on the DGN model for federated CBT learning.
Specifically, as shown in Algorithm~\ref{alg:fedcbt}, a FL of CBT involves two loops for different FL rounds and different hospitals, respectively. Suppose we have $T$ rounds and $K$ hospitals (i.e., sites), we compute the network weights $\mathbf{w}_k^{(t+1)}$ for the $k$th ($k\in\{1, 2, \cdots, K\}$) hospital at the $t$th ($t\in\{1, 2, \cdots, T-1\}$) round using
\begin{equation}
\mathbf{w}_k^{(t+1)} =\texttt{DGN}_k(\mathbf{w}_k^{(t)}, \mathcal{T}_k, E),
\end{equation}
where $\texttt{DGN}_k(\mathbf{w}_k^{(t)}, \mathcal{T}_k, E)$ is the \cbt{} model at $k$th hospital with the network weights of the last round $\mathbf{w}_k^{(t)}$ and is trained over $E$ epochs with the training data $\mathcal{T}_k$ at the $k$th hospital and an subject normalization loss (SNL) function \cite{2020Deep}.

After the network weights are updated for all hospitals, we obtain a set of weights $\{\mathbf{w}_k^{(t+1)}\}_{k=1}^{K}$. We then update the global \cbt{} model by computing the weighted average weights $\mathbf{w}^{(t+1)}$ using
\begin{equation}
\mathbf{w}^{(t+1)}= \sum_{k=1}^K\frac{N_k}{N}\mathbf{w}_k^{(t+1)},
\end{equation}
where $N$ denotes the total number of subjects among all hospitals, while $N_k$ denotes the number of subjects at the $k$th hospital.
After finishing $T$ FL rounds, we have the converged global model for predicting the expected CBT.

\subsubsection{Metadata-Driven Federated CBT learning}
Algorithm~\ref{alg:medfed} shows an overview of \ourmodel{}.
As can be observed, \ourmodel{} differs from conventional federated CBT learning (Algorithm~\ref{alg:fedcbt}) in terms of three aspects.
{First, a pre-federated-learning procedure (Setp 4 and 5) is introduced to construct a set of CBTs $\{\mathbf{\hat{C}}_k\}_{k=1}^{K}$ for generating brain connectivities, which are utilized to train metadata regressors $\{\texttt{MR}_k(\cdot,\cdot,\cdot)\}_{k=1}^{K}$.}
Secondly, the network weight learning at different hospitals is improved by including the federated meta-domain data as follows
\begin{equation}
\mathbf{w}_k^{(t+1)} =\texttt{DGN}_k(\mathbf{w}_k^{(t)}, \mathcal{\hat{T}}_k, E);~\mathcal{\hat{T}}_k = \mathcal{T}_k \bigcup \mathcal{M}_k^{(t+1)}
\end{equation}
where $\mathcal{\hat{T}}_k$ is the new $k$th training data consisting of both original training data $\mathcal{T}_k$ and metadata-driven generated data $\mathcal{M}_k$.
Third, a post-federated-learning procedure is introduced to update the training data by (i) estimating the metadata $\textbf{m}_k$ via $\textbf{m}_k =\texttt{MR}_k(\hat{\mathbf{w}}_k^{(t+1)},\hat{\mathbf{w}}^{(t+1)})$, (ii) generating brain connectivity data via $\mathcal{M}_k = \texttt{MCG}_k(\textbf{m}_k, \textbf{C}_k)$, and (iii) updating the training data via $\mathcal{\hat{T}}_k = \mathcal{T}_k \bigcup \mathcal{M}_k$,
where $\texttt{MCG}_k(\textbf{m}_k, \textbf{C}_k)$ is a metadata-driven connectivity generator with metadata $\textbf{m}_k$ and CBT $\textbf{C}_k$ as inputs.
The DGN model is trained with an enhanced SNL function, which considers samples from both $\mathcal{T}_k$ and $\mathcal{M}_k$.

The metadata $\textbf{m}_k$ captures the underlying statistical attributes of the meta-domain, which provides the key cross-hospital information to address the non-IID issue.
The connectivities generated with the metadata act as vital supplementary information for a hospital to learn the distributions of data from other hospitals without raw data sharing.
\rev{To guide the connectivity generation, we define the metadata $\textbf{m}_k$ as the mean ${\mu}_m$ and standard derivation ${\sigma}_m$ of  distribution of the Gaussian noise added on a centralized subject template, i.e., $\textbf{m}_k = [{\mu}_m,{\sigma}_m]$.}
In what follows, we will detail each component of our \ourmodel{}, including the metadata regressor $\texttt{MR}_k(\cdot,\cdot)$, shown in Fig.~\ref{fig:overview}(c), and the metadata-driven connectivity generator $\texttt{MCG}_k(\cdot,\cdot)$, shown in Fig.~\ref{fig:overview}(b).


\subsection{{Metadata Regressor}}
The metadata regressor $\texttt{MR}_k(\cdot,\cdot)$ plays an important role in our \ourmodel{} since it predicts the metadata for guiding the connectivity generation in each hospital.
To predict the metadata, we build on the fact that the residual network weights (i.e., the subtraction of the local model and global model) encode the underlying statistical attributes of the meta-domain (i.e., metadata).
Therefore, we design our metadata regressor to learn the mapping between residual network weights and metadata.
In our framework, we implement $\texttt{MR}_k(\cdot,\cdot)$ using a multilayer perceptron (MLP) with the network weights of two \cbt{} models as input and the expected metadata parameters as output.
These two parameters determine the statistics of the perturbed CBT, which acts as the center of the metadata.
We train the metadata regressor prior to the FL start.
For this purpose, we randomly generate $N$ sets of brain connectivity multigraphs using our metadata-driven connectivity generator detailed in Section~\ref{section:A}, construct $N$ corresponding CBTs using the data, and train the regressor with paired network weights and metadata parameters.
For simplicity, we omit the hospital index $k$ and introduce the training of metadata regressor at each hospital as follows:
\begin{enumerate}
	\item {Before the first round of FL starts, we randomly generate $M$ sets of brain multigraph data {with our metadata-driven connectivity generator on random Gaussian noise determined by the metadata parameters $\{\textbf{m}_m\}_{m=1}^{M}$}, and the noise matrix of the $m$th metadata is $\mathbf{N}_m \sim \mathcal{N}({\mu_m},{\sigma_m}).$}
	\item {Then, in each round, and for each hospital (i.e., source), we then train a set of \cbt{} models on the source domain $\mathcal{T}_o$ and the generated meta domains $\{\mathcal{M}_m\}_{m=1}^{M}$, each of which is provided by a \cbt{} model trained with local brain multigraphs.}  $\texttt{DGN}_{\text{s}}^{(n)}(\cdot)$ 
	\item {Based on the \cbt{} models, we compute the network weights $\{\textbf{w}_{\text{avg}}^{(m)}\}_{m=1}^{M}= \text{avg}(\textbf{w}^{(m)}+\textbf{w}_{\text{o}})$. The DGN model trained with original data corresponds to the network weights of $\textbf{w}_{\text{o}}$. }
	\item {We train the metadata regressor $\texttt{MR}(\cdot,\cdot)$ with network weights and metadata parameters $\{((\textbf{w}_{\text{avg}}^{(m)},\textbf{w}_{\text{o}}),\textbf{m}_m)\}_{m=1}^{M}$.}
\end{enumerate}
For each hospital, we train a corresponding metadata regressor and obtain a set of regressors $\{\texttt{MR}_{k}(\cdot,\cdot)\}_{k=1}^{K}$, which are used to predict the metadata during our FL procedure, as detailed in Algorithm~\ref{alg:medfed}.

\subsection{Metadata-Driven Connectivity Generator}
\label{section:A}
\begin{figure}[t]
	\centering
 \mybox{{\includegraphics[trim={0cm 9.7cm 19.5cm 0cm},clip,width=0.95\linewidth,page=2]{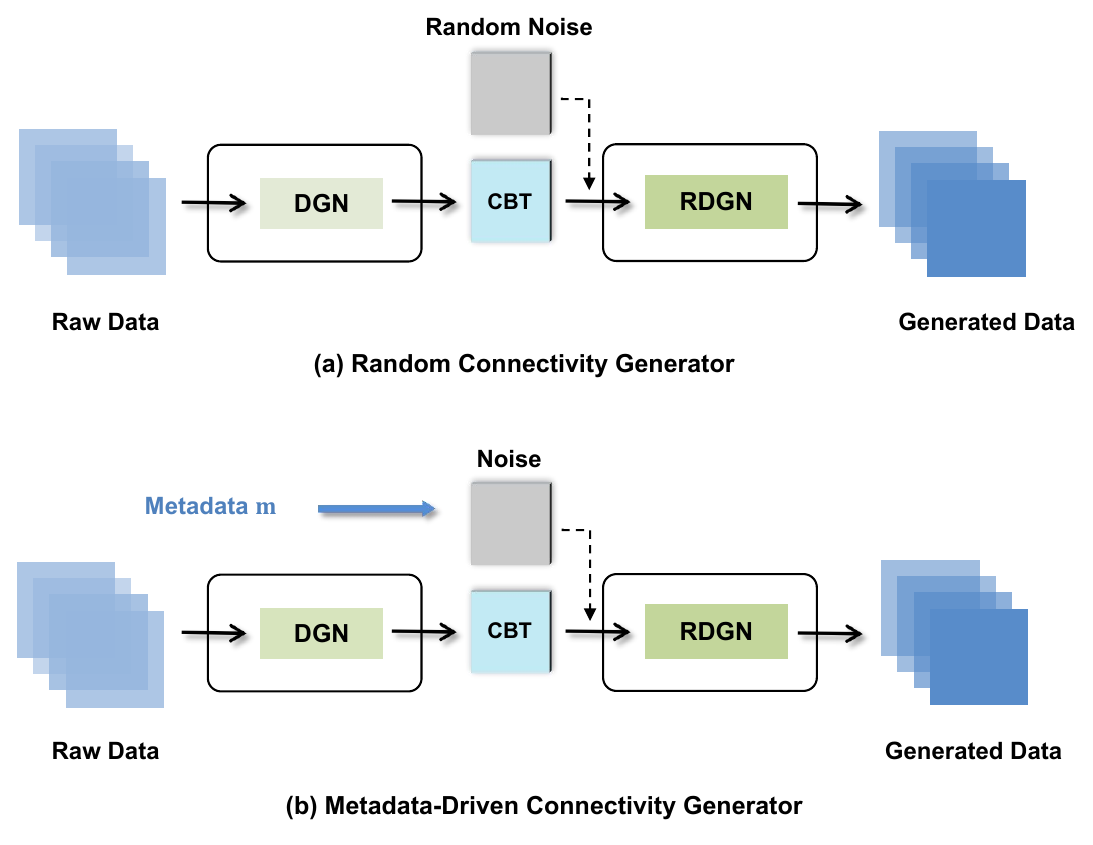}}}
	\vspace{-3pt}
    \caption{{Illustration of our unsupervised (a) and supervised (b) connectivity generation methods.} Through adjusting the distribution parameters of a noisy matrix, our metadata-driven connectivity generator (b) achieves a guided brain connectivity generation.}\label{fig:generator}
\end{figure}
Different from the multigraph generator network \cite{pala2022predicting}, our metadata-driven connectivity generator is able to generate brain multigraphs in a \emph{supervised manner} with guidance from metadata.
We will first detail the overall architecture of our generator and then explain how to leverage the predicted metadata to guide the generation of brain multigraphs.

\subsubsection{Connectivity Generator}
Fig.~\ref{fig:generator}(a)/(b) shows the overall architecture of the unsupervised/supervised version of our connectivity generator.
In general, the generator consists of two components, including a DGN to construct CBT and an reverse DGN \cite{pala2022predicting} to generate connectivity data from the CBT.
The RDGN is based on a U-Net architecture, which is constituted of an encoder, a decoder, and a skip connection.
The encoder is built with four cascaded blocks, each of which consists of a basic unit and a max pooling layer.
The basic unit consists of two convolutional layers with the same kernel size of $3 \times 3$ followed by a ReLU activation function and a batch normalization layer.
The decoder consists of another four cascaded blocks, each of which consists of a transposed convolutional layer with a kernel size of $2 \times 2$ and a basic unit.
{RDGN and DGN are trained in a cyclic and end-to-end manner, ensuring the effective integration and reconstruction of multigraph brain information through the generated CBT.}


As, shown in Fig.~\ref{fig:overview}(a), given a set of training samples with $N$ subjects, the connectivity tensor of subject $n$ is represented by $\mathbf{X}_{n} \in \mathbb{R}^{R \times R \times V}$, where $V$ denotes the number of multi-view brain networks and $R$ denotes the number of graph nodes.
For the generator, we input the training set $\{ \mathbf{X}_{n}\}_{n=1}^{N}$ to DGN model, construct the subject-specific CBTs $\{ \textbf{C}_{n}\}_{n=1}^{N}$, and then generate the brain connectivities $\{\mathbf{X}_{n}^{(m)}\}_{n,m=1}^{N, M}$ with the RDGN model, where $M$ denotes the number of sets of generated brain multigraphs.
The generator is trained with a mixture loss consisting of two terms.
The first term is an SNL function for CBT learning.
The second term minimizes the difference between the predicted connectivities and the raw connectivities used to construct the CBT.


\subsubsection{Metadata-Driven Design}
To achieve supervised connectivity generation, we improve the connectivity generator with metadata.
Specifically, instead of using random noise, as shown in Fig.~\ref{fig:generator}(a), we generate noise according to the metadata, as shown in Fig.~\ref{fig:generator}(b).
For each hospital, we first compute a set of {subject-specific CBTs} from input multigraphs.
{We then generate $M$ set of noise $\{\textbf{N}_{m}\}_{m=1}^{M}$ according to $\mathbf{N}_m \sim \mathcal{N}({\mu}_m,{\sigma}_m)$.}
Finally, the noise $\textbf{N}_{m}$ is added with CBTs $\textbf{C}_{n}$ for the input of RDGN $\texttt{RDGN}(\cdot)$ to generate the connectivities $\textbf{X}_n^{(m)}$.
Mathematically, we define the generated connectivities as
\begin{equation}
    {\textbf{X}}_n^{(m)} = \texttt{RDGN}(\textbf{C}_{n} + \mathbf{N}_m).
\end{equation}
Thanks to the supervised connectivity generation empowered by metadata, the generated multigraphs are nested in an underlying meta-domain, which provides holistic information to the federated CBT learning for addressing the non-IID issue.



\section{Experiments \& Discussion}
\label{section 4}
\subsection{Dataset}
We evaluate our \ourmodel{} with the morphological brain networks of 186 normal control (NC) and 155 autism spectrum disorder (ASD) subjects from autism brain imaging data exchange (ABIDE-I) public dataset \cite{0The}.
Using FreeSurfer \cite{2012FreeSurfer}, left hemisphere (LH) and right hemisphere (RH) cortical surfaces are generated from T1-weighted MR images \cite{nebli2020gender}.
Subsequently, using the morphology of the cortical surface, we construct 6 morphological brain networks using the Desikan-Killiany Atlas \cite{2006An}, including 35 regions of interest (ROIs).
To measure the correlation between two regions, we compute the absolute difference of the average cortical attribute values in these two regions \cite{Mahjoub2018BrainMR}.
The morphological brain networks quantify the shape relationship between different cortical regions for a specific cortical attribute.
For each subject, we have a connectivity tensor constructed from 6 cortical morphological brain networks regarding different surface attributes, including cortical surface area, minimum principle area, maximum principal curvature, average curvature, mean cortical thickness, and mean sulcal depth \cite{mahjoub2018brain,corps2019morphological,bilgen2020machine}.

{We perform experiments using four datasets, including ASD-RH, NC-RH, ASD-LH, and NC-LH.
To simulate the multi-site scenario, we utilize $k$-means \cite{na2010research} to cluster each dataset into three clusters with balanced distributed samples.}
{Next, we split each cluster into four parts to allow four-fold cross-validation.}
In each fold, we take three parts from each cluster to construct the training data, and the rest one is utilized for testing.
We set the fraction of hospitals participating in the weights communication with the server to 0.6.

\subsection{Implementation Details}
Our \ourmodel{} is implemented in PyTorch and PyTorch-Geometric \cite{2019Fast} libraries.
The local/global CBT models in our \ourmodel{} are all based on the DGN \cite{2020Deep}, which learns the CBT from multi-view multi-domain brain connectivities (i.e., muligraphs). 
We train all the models using gradient descent with Adam optimizer \cite{chaari2023comparative} and a learning rate of 0.0008.
In our SNL function \cite{2020Deep}, we fix the proportion of the random samples in each hospital to 0.4.
For the training settings, we train all strategies using $T=500$ rounds with an early stop mechanism of 11 rounds and running 2 epochs in each round.
We are able to execute an update once per epoch without encountering a GPU memory bottleneck, by utilizing the gradients computed for the entire dataset.
For supervised \ourmodel{}, we tuned the parameter of epochs within the first round with the consideration of the stability of training.
To do so, we trained the metadata regressor at 40 epochs ending the first round during the pre-federated-learning step.
Our regressor is trained with an Adam optimizer \cite{chaari2023comparative} with a learning rate of 0.0005.
The brain multigraphs used to train metadata regressor are simulated with our metadata-driven connectivity generator with random metadata values.


\begin{figure*}[t]
	\centering
    \mybox{{\includegraphics[trim={0cm 11cm 20cm 0cm},clip,width=0.99\textwidth,page=2]{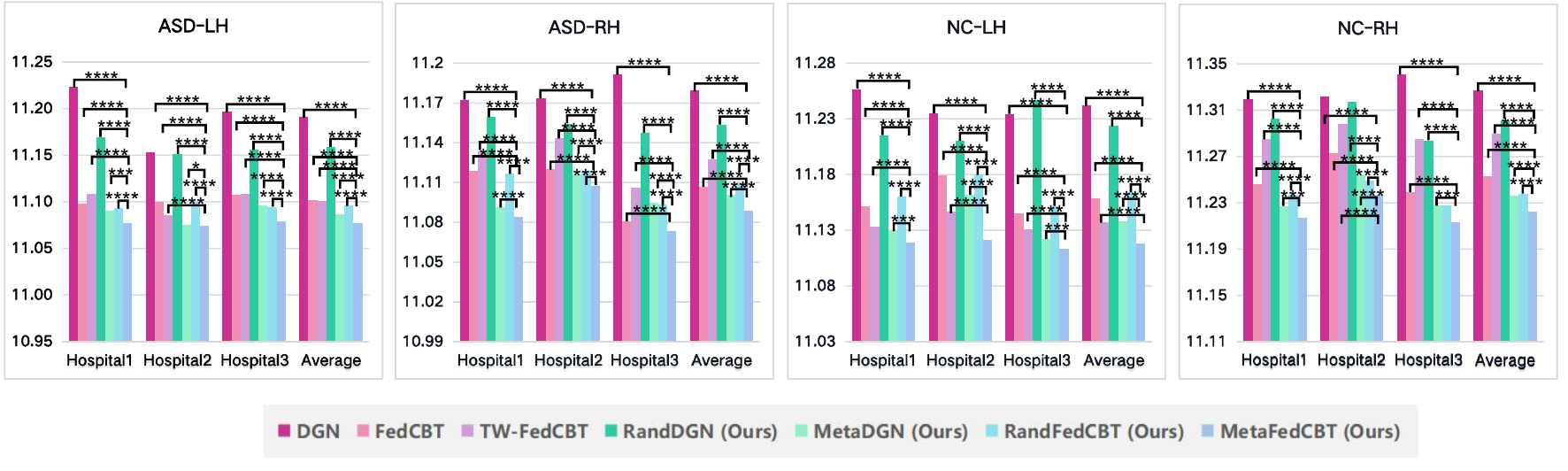}}} 
    \vspace{-6pt}
	\caption{Quantitative comparison of the centeredness of CBTs given by \ourmodel{}, the comparison models, and the ablated versions.
	The averaged Frobenius distance of all folds is adopted for evaluation.
	We also report the $p$-values using a two-tailed paired t-test between \ourmodel{} and each comparison method.
	If the $p$-value is less than 0.05/0.01/0.001/0.0001, it is flagged with one/two/three/four stars (*/**/***/****), respectively.}
	\label{fig:numerical_results} 
\end{figure*}

\begin{figure*}[t]
	\centering
        \mybox{{\includegraphics[trim={0 5cm 20.5cm 0cm},clip,width=0.99\textwidth,page=1]{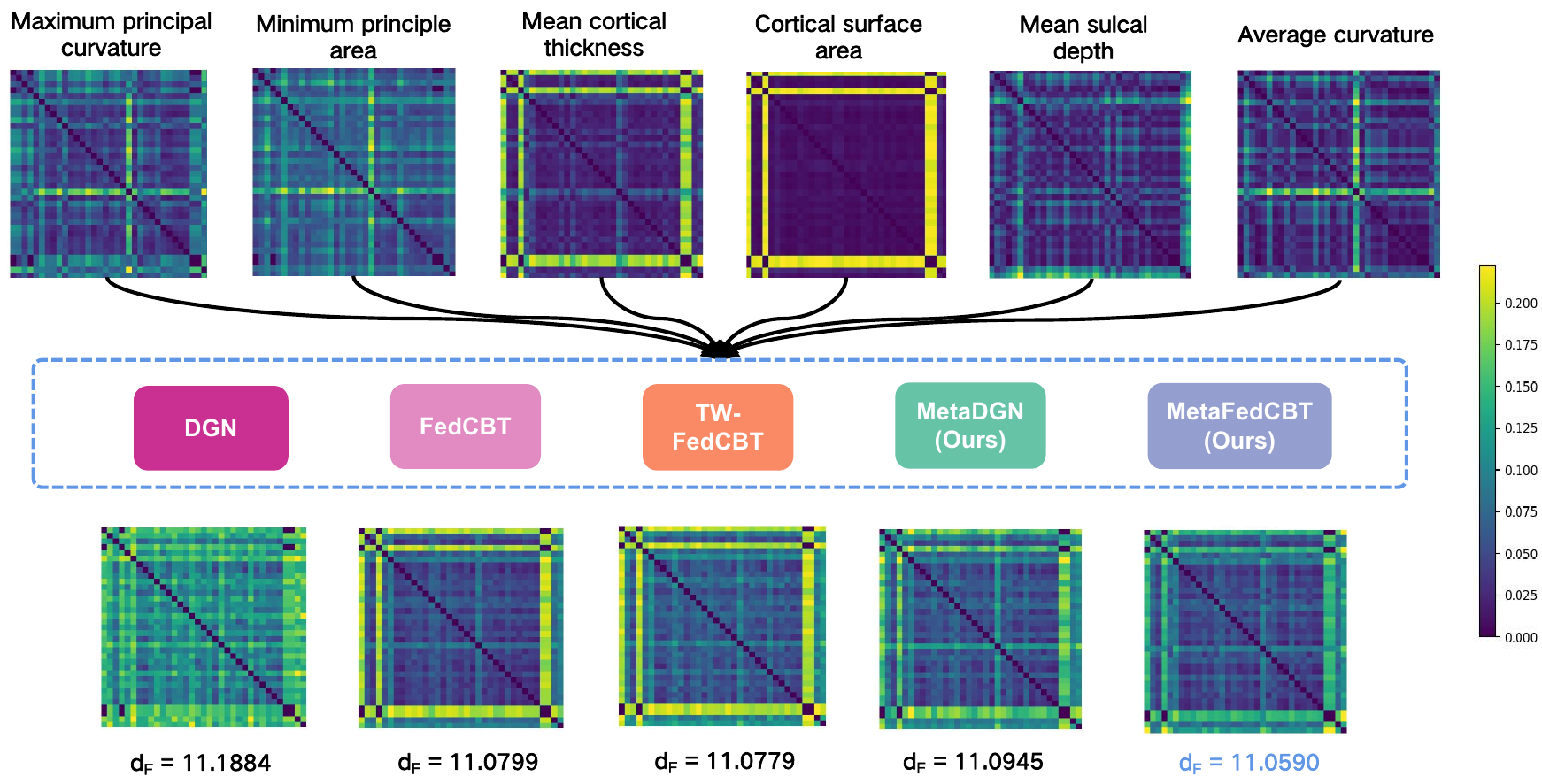}}} 
        \vspace{-5pt}
 \caption{Visual comparison of CBTs given by different models. Experiment was performed on the fold one of ASD-LH dataset.}
	\label{fig:visual} 
\end{figure*}

\begin{figure*}[t]
	\centering
        \mybox{{\includegraphics[trim={0 10cm 12cm 0cm},clip,width=0.99\textwidth,page=4]{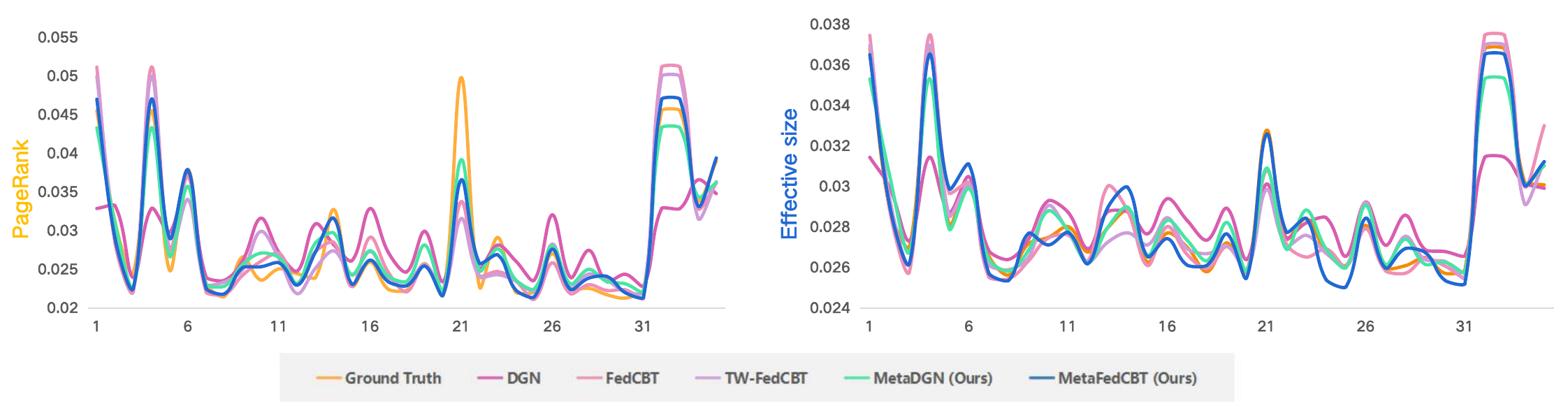}}} 
        \vspace{-6pt}
\caption{Visual comparison of topological distributions of CBTs given by different models. Experiment was performed on the fold one of ASD-LH dataset.}
	\label{fig:visual2} 
\end{figure*}

\begin{figure*}[t]
\centering
\mybox{{\includegraphics[trim={0 2.2cm 10cm 0cm},clip,width=0.99\textwidth,page=5]{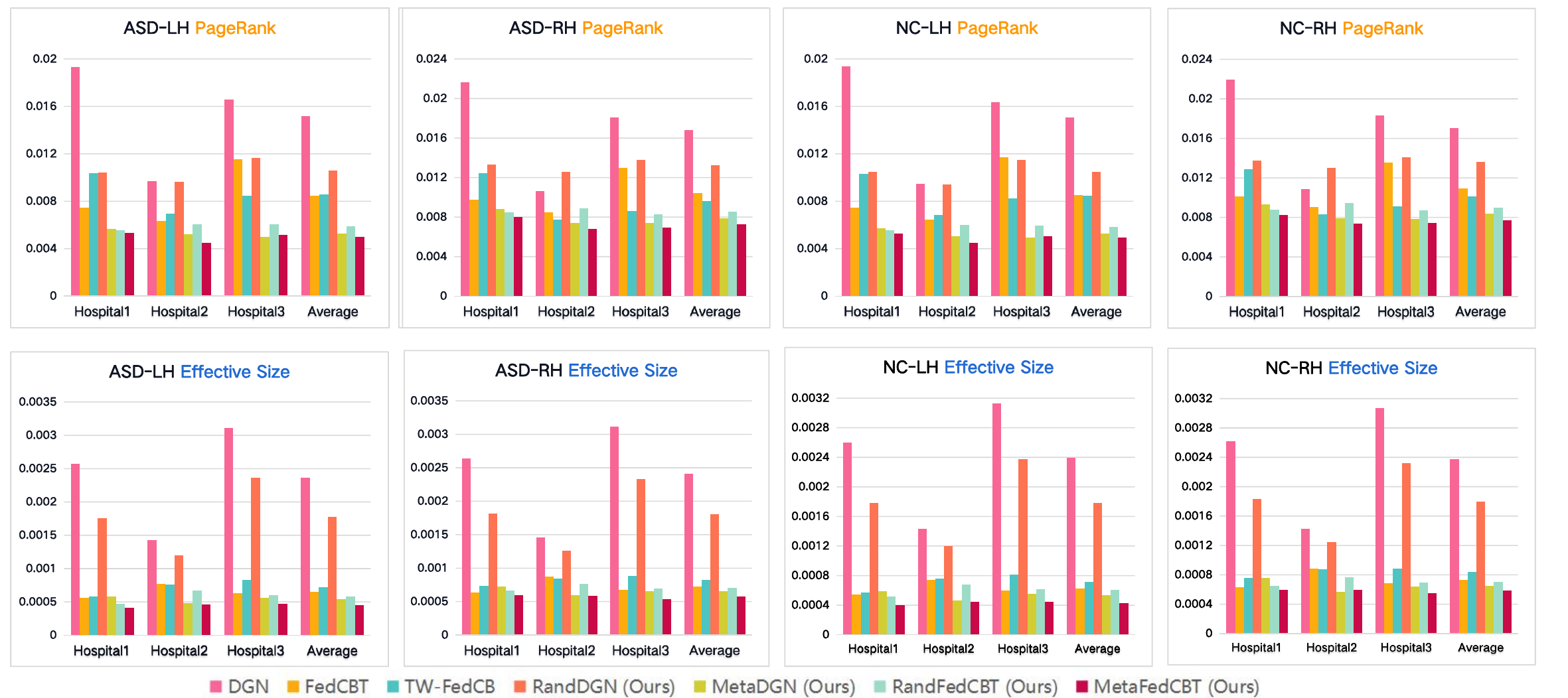}}} 
\vspace{-3pt}
\caption{Quantitative comparison of the topology soundness of CBTs given by different models.
We report the averaged result of all folds for each dataset and each topological measure.}
\label{fig:numerical_results2} 
\end{figure*}
\begin{table*}[ht!]
	\renewcommand{\arraystretch}{1.4}
	\setlength\tabcolsep{4pt}
	\centering
	\caption{Testing classification results of SVM classifiers trained with a single CBT from each class of each fold of each hospital. We report the accuracy, precision, recall, and F1 score of two classification tasks (i.e., NC-LH vs. ASD-LH and NC-RH vs. ASD RH) over the four cross-validation folds across three hospitals.}
	\resizebox{\linewidth}{!}{
		\belowrulesep=0pt
		\aboverulesep=0pt
		\begin{tabular}{p{0.1cm}|c|cccc|cccc|cccc|cccc}
			\toprule[2pt]
			\multicolumn{1}{c}{}  & \multicolumn{1}{c|}{} & \multicolumn{4}{c|}{\textbf{DGN-CBT}} & \multicolumn{4}{c|}{\textbf{FedCBT-CBT}} & \multicolumn{4}{c|}{\textbf{TW-FedCBT}} & \multicolumn{4}{c}{\textbf{MetaFed-CBT (Ours)}} \\
			\cmidrule{3-18}    
			 \multicolumn{1}{c}{} & \multicolumn{1}{c|}{} & \multicolumn{1}{c}{Acc} & \multicolumn{1}{c}{Prec} & \multicolumn{1}{c}{Rec} & \multicolumn{1}{c|}{F1} & \multicolumn{1}{c}{Acc} & \multicolumn{1}{c}{Prec} & \multicolumn{1}{c}{Rec} & \multicolumn{1}{c|}{F1} & \multicolumn{1}{c}{Acc} & \multicolumn{1}{c}{Prec} & \multicolumn{1}{c}{Rec} & \multicolumn{1}{c|}{F1} & \multicolumn{1}{c}{Acc} & \multicolumn{1}{c}{Prec} & \multicolumn{1}{c}{Rec} & \multicolumn{1}{c}{F1} \\
			\midrule[1.5pt]
			\multicolumn{1}{c|}{\multirow{3}[3]{*}{LH}} & \multicolumn{1}{c|}{\multirow{1}[1]{*}{Hospital 1}} 
			& 0.571  & 0.576  & 0.514  & 0.500  & 0.580  & 0.595  & 0.502  & 0.531  & 0.685  & 0.756  & 0.534  & 0.623  & $\pmb{0.880}$ & $\pmb{0.870}$ & $\pmb{0.900}$ & $\pmb{0.883}$ \\
   & \multicolumn{1}{c|}
   {\multirow{1}[1]{*}{Hospital 2}} 
   & 0.668  & 0.617  & 0.883  & 0.726  & 0.710  & 0.740  & 0.714  & 0.712  & 0.618  & 0.748  & 0.339  & 0.464  & $\pmb{0.929}$ & $\pmb{0.921}$ & $\pmb{0.942}$ & $\pmb{0.929}$ \\
   & \multicolumn{1}{c|}
   {\multirow{1}[1]{*}{Hospital 3}} 
   & 0.531  & 0.536  & 0.562  & 0.531  & 0.616  & 0.602  & 0.581  & 0.588  & 0.586  & 0.579  & 0.637  & 0.602  & $\pmb{0.913}$ & $\pmb{0.945}$ & $\pmb{0.884}$ & $\pmb{0.910}$ \\
			\midrule[1.5pt]
			\multicolumn{1}{c|}{\multirow{3}[3]{*}{RH}} & \multicolumn{1}{c|}{\multirow{1}[1]{*}{Hospital 1}} 
   & 0.633  & 0.584  & 0.603  & 0.571  & 0.910  & $\pmb{0.905}$ & 0.892  & 0.897  & 0.599  & 0.531  & 0.544  & 0.527  & $\pmb{0.915}$ & 0.884  & $\pmb{0.935}$ & $\pmb{0.907}$ \\
   & \multicolumn{1}{c|}{\multirow{1}[1]{*}{Hospital 2}} 
   & 0.653  & 0.577  & 0.718  & 0.646  & 0.683  & 0.658  & 0.542  & 0.592  & 0.531  & 0.498  & 0.562  & 0.515  & $\pmb{0.805}$ & $\pmb{0.857}$& $\pmb{0.791}$& $\pmb{0.824}$ \\
   & \multicolumn{1}{c|}{\multirow{1}[1]{*}{Hospital 3}} 
   & 0.606  & 0.588  & 0.473  & 0.487  & 0.709  & 0.644  & 0.737  & 0.687  & 0.498  & 0.450  & 0.778  & 0.563  & $\pmb{0.819}$ & $\pmb{0.783}$ & $\pmb{0.810}$ & $\pmb{0.795}$ \\
			\bottomrule[2pt]
		\end{tabular}%
		\label{tab:classification_results}}
\end{table*}%
\subsection{Results}
{We train different FL models to generate CBTs for the RH and LH of both NC and ASD populations, respectively.}
The resulting CBTs are evaluated in terms of centeredness, topological soundness, and discriminativeness.
We consider their state-of-the-art (SOTA) models, DGN \cite{2020Deep}, FedCBT \cite{2021A}, and TW-FedCBT \cite{2021A}, as benchmarks for basis comparison.
To further investigate the effectiveness of \ourmodel{}, we further introduce three ablated versions on the selection of federation and the supervised impact of metadata.
The ablated versions include
\begin{enumerate}
    \item \textbf{RandDGN}: \ourmodel{} without FL and our metadata-driven connectivity generator, i.e., DGN model with a random connectivity generator;
    \item \textbf{MetaDGN:} \ourmodel{} without FL, i.e., DGN model with our metadata-driven connectivity generator;
    \item \textbf{RandFedCBT:} \ourmodel{} without the metadata-driven connectivity generator, i.e., FedCBT with a random connectivity generator.
\end{enumerate}
For the unsupervised ablated version of our \ourmodel{}, we perform random data simulation multiple times (10 in our case) for more stable results.

An ideal CBT constructed by different models is expected to possess three key characteristics.
First, it should be well-centered, meaning that it satisfies the minimum distance requirement for all subjects in the population.
Second, it should be discriminative, capturing the unique population connectivities.
Last, it should be topologically sound, preserving globally the topological properties of the individual complex brain networks after being mapped into a population-driven template.
\subsubsection{CBT Centeredness Test}
{We measure the average Frobenius distance between the learned CBT and the tensor view of each subject from the training set to evaluate the centeredness of the constructed global CBT.}
The Frobenius distance between two matrices, $\mathbf{A}$ and $\mathbf{B}$, is computed as: $d_{F}(\mathbf{A}, \mathbf{B})=\sqrt{\sum_{i} \sum_{j}{\left|\mathbf{A}_{i,j}-\mathbf{B}_{i,j}\right|}^{2}}.$
By averaging the mean Frobenius distance from the estimated CBT to all views of a connectivity tensor of each subject from the testing set, we assess the centeredness of the estimated CBTs given by different comparison methods and ablated versions across all folds of each hospital for all four datasets, as shown in Fig.~\ref{fig:numerical_results}.
The results indicate that our \ourmodel{} provides the best performance with the minimum Frobenius distance across all datasets and hospitals.
This demonstrates that our \ourmodel{} is an effective federated CBT learning model and significantly (p\textless 0.001) outperforms the SOTA comparison models in terms of centeredness across all folds and evaluation datasets.
Furthermore, the full version of \ourmodel{} also outperforms its ablated versions (i.e., RandDGN, MetaDGN, and RandFed), indicating that metadata-driven connectivity generation and FL are effective in improving the CBT learning from non-IID multi-site connectivities.
Especially, the supervised versions (i.e., MetaDGN and \ourmodel{}) show an obvious improvement against the unsupervised versions (i.e., RandDGN and RandFedCBT).
It shows that our metadata regressor effectively conducts the generation and can capture valid statistical information from other hospitals' data. With the local model updating of FL based on MetaDGN, \ourmodel{} slightly outperforms the simpler version in many folds and datasets, which can be explained by the fact that our method needs a round-by-round model weights communication to correct the bias of predicted metadata by the regressor in some cases.

{We also show the visual comparison of CBTs of hospital one in Fig.~\ref{fig:visual}.} As can be observed, our \ourmodel{} and MetaDGN generate the CBTs with the smallest $d_{F}$ values of the ASD-RH dataset, indicating the best performance in CBT-centeredness.
The visual results, shown in Fig.~\ref{fig:visual}, further confirm our observations in Fig.~\ref{fig:numerical_results}.

\subsubsection{CBT Topological Soundness Test}
We conduct an assessment of the topological validity of the learned CBTs at the node-wise scale. We use the different topological measures to capture graph topological properties of brain connectivity, including PageRank \cite{brin1998pagerank} and effective node size \cite{lazega1995structural}.
For each measure, a discrete probability distribution can be derived by calculating the topological measures for each node and then normalizing them with respect to the total sum of measures across all nodes. In order to have a comprehensive topological analysis, we first average the distribution of topological measures of multi-view connectivity graphs of each testing subject as ground truth. We then use the Kullback-Leibler (KL) divergence measure to compare the estimated CBT constructed from the training dataset with this ground truth.
Especially, the KL divergence of the ground truth distributions $p$ and distributions $q$ derived from the CBT is defined as $D_{\text{KL}}(p \| q)=\sum_{i=1}^{N} p\left(x_{i}\right)\left(\log p\left(x_{i}\right)-\log q\left(x_{i}\right)\right)$.

We show the visual result of the cross-node distribution of fold one of the ASD-LH dataset in Fig.~\ref{fig:visual2} and KL divergence results compared with the averaged ground truth of PageRank and effective size measured for the ASD-LH in Fig.~\ref{fig:numerical_results2}.
As can be observed, the CBT of our \ourmodel{} is most similar to the group represented by multi-view brain networks in encapsulating complex topological patterns.
Our \ourmodel{} outperforms both SOTA comparison methods and its ablated versions, which sufficiently shows the superiority of \ourmodel{} and the effectiveness of the proposed components in terms of topology soundness.


\subsubsection{CBT Discriminativeness Test}
A representative CBT can capture the unique characteristics of brain networks of a given population, making it easily distinguishable from the CBT of another population.
These distinctive features can be utilized in clinical applications for biological classification tasks, such as distinguishing informative biomarkers to facilitate disease diagnosis.
To demonstrate the discriminative ability of our federated CBT, we perform an additional evaluation, which trains an SVM classifier using one-shot learning, with two CBTs, one for the ASD population and the other for the NC population, as input.
The trained SVM is employed to test with the input of the ASD and NC subject-specific CBTs from the global testing set.
We set up two classification tasks, namely ASD-LH vs. NC-LH and ASD-RH vs. NC-RH.
Table~\ref{tab:classification_results} reports the comparison of classification accuracy, precision, recall, and F1 scores between different methods. 
The results indicate that CBTs learned by our \ourmodel{} are most discriminative compared with SOTA methods, where they yield the highest classification performance.
This further confirms our observations in Fig.~\ref{fig:numerical_results},~\ref{fig:visual},~\ref{fig:numerical_results2},~and~\ref{fig:visual2}.

Our extensive experimental results showed that our \ourmodel{} outperforms comparison models, including FedCBT, and its ablated versions in all three criteria.
This suggests two conclusions.
First, generated multi-domain data can improve the performance of both conventional CBT learning and federated CBT learning.
Second, metadata can further improve performance with effective supervised learning, which boosts the quality of the generated multidomain data.


\section{Conclusion}
\label{section 5}
In this work, we proposed \ourmodel{}, the first FL framework for CBT learning from disparate brain connectivity domains.
Our \ourmodel{} overcomes the limitations of existing methods that are unable to handle statistical data heterogeneity, (i.e., data/feature shift due to disparate hospital-specific connectivity patterns and distributions).
It is able to predict metadata that approximates other local domains in a fully supervised manner and leverages the metadata to generate effective connectivity data for boosting model's robustness to non-IID data and CBT representativeness.
Our results demonstrate the effectiveness of \ourmodel{} in privacy-preserving CBT learning using connectivity datasets.

\bibliographystyle{IEEEtran}
\bibliography{references}

\begin{thebibliography}{10}
\providecommand{\url}[1]{#1}
\csname url@samestyle\endcsname
\providecommand{\newblock}{\relax}
\providecommand{\bibinfo}[2]{#2}
\providecommand{\BIBentrySTDinterwordspacing}{\spaceskip=0pt\relax}
\providecommand{\BIBentryALTinterwordstretchfactor}{4}
\providecommand{\BIBentryALTinterwordspacing}{\spaceskip=\fontdimen2\font plus
\BIBentryALTinterwordstretchfactor\fontdimen3\font minus
  \fontdimen4\font\relax}
\providecommand{\BIBforeignlanguage}[2]{{%
\expandafter\ifx\csname l@#1\endcsname\relax
\typeout{** WARNING: IEEEtran.bst: No hyphenation pattern has been}%
\typeout{** loaded for the language `#1'. Using the pattern for}%
\typeout{** the default language instead.}%
\else
\language=\csname l@#1\endcsname
\fi
#2}}
\providecommand{\BIBdecl}{\relax}
\BIBdecl

\bibitem{2019Estimation}
S.~Dhifallah and I.~Rekik, ``Estimation of connectional brain templates using
  selective multi-view network normalization,'' \emph{Medical Image Analysis},
  vol.~59, p. 101567, 2019.

\bibitem{chaari2023comparative}
N.~Chaari, H.~C. Akda{\u{g}}, and I.~Rekik, ``Comparative survey of multigraph
  integration methods for holistic brain connectivity mapping,'' \emph{Medical
  Image Analysis}, p. 102741, 2023.

\bibitem{2019Clustering}
S.~Dhifallah, I.~Rekik, A.~D.~N. Initiative \emph{et~al.}, ``Clustering-based
  multi-view network fusion for estimating brain network atlases of healthy and
  disordered populations,'' \emph{Journal of neuroscience methods}, vol. 311,
  pp. 426--435, 2019.

\bibitem{2020Estimation}
S.~Dhifallah, I.~Rekik, and A.~D.~N. Initiative, ``Estimation of connectional
  brain templates using selective multi-view network normalization.''
  \emph{Elsevier}, 2020.

\bibitem{2020Deep}
M.~B. Gurbuz and I.~Rekik, ``Deep graph normalizer: a geometric deep learning
  approach for estimating connectional brain templates,'' in
  \emph{MICCAI}.\hskip 1em plus 0.5em minus 0.4em\relax Springer, 2020, pp.
  155--165.

\bibitem{2021MGN}
I.~Rekik and M.~B. Gurbuz, ``Mgn-net: a multi-view graph normalizer for
  integrating heterogeneous biological network populations,'' \emph{Medical
  Image Analysis}, 2021.

\bibitem{2020FedNER}
S.~Ge, F.~Wu, C.~Wu, T.~Qi, and X.~Xie, ``Fedner: Medical named entity
  recognition with federated learning,'' 2020.

\bibitem{FedAvg}
B.~McMahan, E.~Moore, D.~Ramage, S.~Hampson, and B.~A. y~Arcas,
  ``Communication-efficient learning of deep networks from decentralized
  data,'' in \emph{Artificial intelligence and statistics}.\hskip 1em plus
  0.5em minus 0.4em\relax PMLR, 2017, pp. 1273--1282.

\bibitem{2021A}
H.~C. Bayram and I.~Rekik, ``A federated multigraph integration approach for
  connectional brain template learning,'' in \emph{Multimodal Learning for
  Clinical Decision Support: 11th International Workshop}.\hskip 1em plus 0.5em
  minus 0.4em\relax Springer, 2021, pp. 36--47.

\bibitem{li2020federated}
T.~Li, A.~K. Sahu, A.~Talwalkar, and V.~Smith, ``Federated learning:
  Challenges, methods, and future directions,'' \emph{IEEE signal processing
  magazine}, vol.~37, no.~3, pp. 50--60, 2020.

\bibitem{10077569}
M.~F. Sohan and A.~Basalamah, ``A systematic review on federated learning in
  medical image analysis,'' \emph{IEEE Access}, vol.~11, pp. 28\,628--28\,644,
  2023.

\bibitem{sarma2021federated}
K.~V. Sarma, S.~Harmon, T.~Sanford, H.~R. Roth, Z.~Xu, J.~Tetreault, D.~Xu,
  M.~G. Flores, A.~G. Raman, R.~Kulkarni \emph{et~al.}, ``Federated learning
  improves site performance in multicenter deep learning without data
  sharing,'' \emph{Journal of the American Medical Informatics Association},
  vol.~28, no.~6, pp. 1259--1264, 2021.

\bibitem{Adnan2021FederatedLA}
M.~Adnan, S.~Kalra, J.~C. Cresswell, G.~W. Taylor, and H.~R. Tizhoosh,
  ``Federated learning and differential privacy for medical image analysis,''
  \emph{Scientific Reports}, vol.~12, 2021.

\bibitem{pati2022federated}
S.~Pati, U.~Baid, B.~Edwards, M.~Sheller, S.-H. Wang, G.~A. Reina, P.~Foley,
  A.~Gruzdev, D.~Karkada, C.~Davatzikos \emph{et~al.}, ``Federated learning
  enables big data for rare cancer boundary detection,'' \emph{Nature
  communications}, vol.~13, no.~1, p. 7346, 2022.

\bibitem{nguyen2021federated}
D.~C. Nguyen, M.~Ding, P.~N. Pathirana, A.~Seneviratne, and A.~Y. Zomaya,
  ``Federated learning for covid-19 detection with generative adversarial
  networks in edge cloud computing,'' \emph{IEEE Internet of Things Journal},
  vol.~9, no.~12, pp. 10\,257--10\,271, 2021.

\bibitem{nalawade2021federated}
S.~Nalawade, C.~Ganesh, B.~Wagner, D.~Reddy, Y.~Das, F.~F. Yu, B.~Fei, A.~J.
  Madhuranthakam, and J.~A. Maldjian, ``Federated learning for brain tumor
  segmentation using mri and transformers,'' in \emph{International MICCAI
  Brainlesion Workshop}.\hskip 1em plus 0.5em minus 0.4em\relax Springer, 2021,
  pp. 444--454.

\bibitem{liu2021feddg}
Q.~Liu, C.~Chen, J.~Qin, Q.~Dou, and P.-A. Heng, ``Feddg: Federated domain
  generalization on medical image segmentation via episodic learning in
  continuous frequency space,'' in \emph{Proceedings of the IEEE/CVF Conference
  on Computer Vision and Pattern Recognition}, 2021, pp. 1013--1023.

\bibitem{yang2021federated}
D.~Yang, Z.~Xu, W.~Li, A.~Myronenko, H.~R. Roth, S.~Harmon, S.~Xu, B.~Turkbey,
  E.~Turkbey, X.~Wang \emph{et~al.}, ``Federated semi-supervised learning for
  covid region segmentation in chest ct using multi-national data from china,
  italy, japan,'' \emph{Medical image analysis}, vol.~70, p. 101992, 2021.

\bibitem{yan2020variation}
Z.~Yan, J.~Wicaksana, Z.~Wang, X.~Yang, and K.-T. Cheng, ``Variation-aware
  federated learning with multi-source decentralized medical image data,''
  \emph{IEEE Journal of Biomedical and Health Informatics}, vol.~25, no.~7, pp.
  2615--2628, 2020.

\bibitem{liu2021federated}
Q.~Liu, H.~Yang, Q.~Dou, and P.-A. Heng, ``Federated semi-supervised medical
  image classification via inter-client relation matching,'' in
  \emph{MICCAI}.\hskip 1em plus 0.5em minus 0.4em\relax Springer, 2021, pp.
  325--335.

\bibitem{luo2022fedsld}
J.~Luo and S.~Wu, ``Fedsld: Federated learning with shared label distribution
  for medical image classification,'' in \emph{ISBI}.\hskip 1em plus 0.5em
  minus 0.4em\relax IEEE, 2022, pp. 1--5.

\bibitem{gurler2022federated}
Z.~G{\"u}rler and I.~Rekik, ``Federated brain graph evolution prediction using
  decentralized connectivity datasets with temporally-varying acquisitions,''
  \emph{IEEE Transactions on Medical Imaging}, 2022.

\bibitem{sohan2023systematic}
M.~F. Sohan and A.~Basalamah, ``A systematic review on federated learning in
  medical image analysis,'' \emph{IEEE Access}, 2023.

\bibitem{li2021fedbn}
X.~Li, M.~Jiang, X.~Zhang, M.~Kamp, and Q.~Dou, ``Fedbn: Federated learning on
  non-iid features via local batch normalization,'' \emph{arXiv preprint
  arXiv:2102.07623}, 2021.

\bibitem{sun2021partialfed}
B.~Sun, H.~Huo, Y.~Yang, and B.~Bai, ``Partialfed: Cross-domain personalized
  federated learning via partial initialization,'' \emph{Advances in Neural
  Information Processing Systems}, vol.~34, pp. 23\,309--23\,320, 2021.

\bibitem{wicaksana2022customized}
J.~Wicaksana, Z.~Yan, X.~Yang, Y.~Liu, L.~Fan, and K.-T. Cheng, ``Customized
  federated learning for multi-source decentralized medical image
  classification,'' \emph{IEEE Journal of Biomedical and Health Informatics},
  vol.~26, no.~11, pp. 5596--5607, 2022.

\bibitem{pala2022predicting}
F.~Pala and I.~Rekik, ``Predicting brain multigraph population from a single
  graph template for boosting one-shot classification,'' in \emph{Predictive
  Intelligence in Medicine: 5th International Workshop, PRIME 2022, Held in
  Conjunction with MICCAI 2022, Singapore, September 22, 2022, Proceedings},
  vol. 13564.\hskip 1em plus 0.5em minus 0.4em\relax Springer Nature, 2022, p.
  191.

\bibitem{0The}
A.~Di~Martino, C.-G. Yan, Q.~Li, E.~Denio, F.~X. Castellanos, K.~Alaerts, J.~S.
  Anderson, M.~Assaf, S.~Y. Bookheimer, M.~Dapretto \emph{et~al.}, ``The autism
  brain imaging data exchange: towards a large-scale evaluation of the
  intrinsic brain architecture in autism,'' \emph{Molecular psychiatry},
  vol.~19, no.~6, pp. 659--667, 2014.

\bibitem{2012FreeSurfer}
B.~Fischl, ``Freesurfer,'' \emph{NeuroImage}, vol.~62, no.~2, pp. 774--781,
  2012.

\bibitem{nebli2020gender}
A.~Nebli and I.~Rekik, ``Gender differences in cortical morphological
  networks,'' \emph{Brain imaging and behavior}, vol.~14, no.~5, pp.
  1831--1839, 2020.

\bibitem{2006An}
R.~S. Desikan, F.~S{\'e}gonne, B.~Fischl, B.~T. Quinn, B.~C. Dickerson,
  D.~Blacker, R.~L. Buckner, A.~M. Dale, R.~P. Maguire, and B.~T.~a. Hyman,
  ``An automated labeling system for subdividing the human cerebral cortex on
  mri scans into gyral based regions of interest.'' \emph{NeuroImage}, vol.~31,
  no.~3, pp. 968--980, 2006.

\bibitem{Mahjoub2018BrainMR}
I.~Mahjoub, M.~Mahjoub, and I.~Rekik, ``Brain multiplexes reveal morphological
  connectional biomarkers fingerprinting late brain dementia states,''
  \emph{Scientific Reports}, vol.~8, 2018.

\bibitem{mahjoub2018brain}
I.~Mahjoub, M.~A. Mahjoub, and I.~Rekik, ``Brain multiplexes reveal
  morphological connectional biomarkers fingerprinting late brain dementia
  states,'' \emph{Scientific reports}, vol.~8, no.~1, pp. 1--14, 2018.

\bibitem{corps2019morphological}
J.~Corps and I.~Rekik, ``Morphological brain age prediction using multi-view
  brain networks derived from cortical morphology in healthy and disordered
  participants,'' \emph{Scientific reports}, vol.~9, no.~1, p. 9676, 2019.

\bibitem{bilgen2020machine}
I.~Bilgen, G.~Guvercin, and I.~Rekik, ``Machine learning methods for brain
  network classification: application to autism diagnosis using cortical
  morphological networks,'' \emph{Journal of neuroscience methods}, vol. 343,
  p. 108799, 2020.

\bibitem{na2010research}
S.~Na, L.~Xumin, and G.~Yong, ``Research on k-means clustering algorithm: An
  improved k-means clustering algorithm,'' in \emph{2010 Third International
  Symposium on intelligent information technology and security
  informatics}.\hskip 1em plus 0.5em minus 0.4em\relax Ieee, 2010, pp. 63--67.

\bibitem{2019Fast}
M.~Fey and J.~E. Lenssen, ``Fast graph representation learning with pytorch
  geometric,'' 2019.

\bibitem{brin1998pagerank}
S.~Brin, ``The pagerank citation ranking: bringing order to the web,''
  \emph{Proceedings of ASIS, 1998}, vol.~98, pp. 161--172, 1998.

\bibitem{lazega1995structural}
E.~Lazega, ``Structural holes: the social structure of competition,'' 1995.

\end{thebibliography}

\end{document}


\title{Supplementary Materials}

\maketitle

\begin{figure}[h]
	\centering
        \includegraphics[trim={0 10.5cm 10cm 0cm},clip,width=0.98\textwidth,page=2]{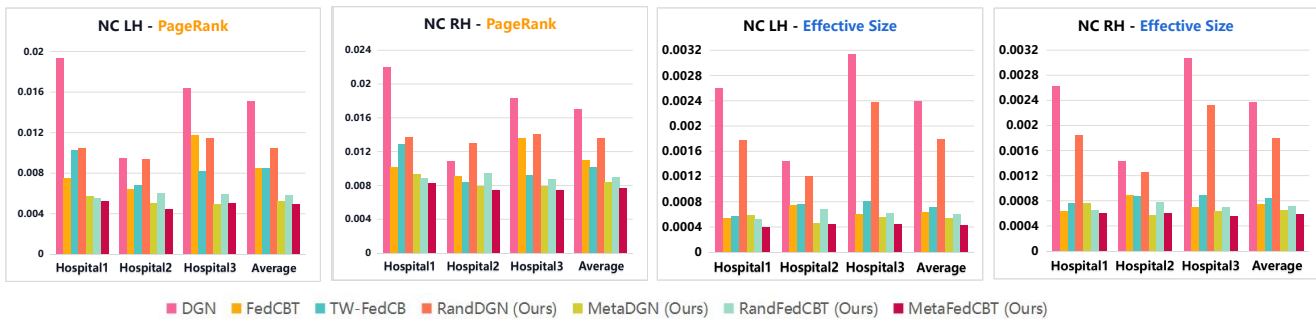} 
        \vspace{-3pt}
 \caption{Quantitative comparison of the topology soundness of CBTs given by \ourmodel{}, the comparison models, and the ablated versions. It is evaluated by the Kullback Liebler divergence of the constructed connectional templates and the ground truth for the PageRank and effective size distributions. We report the averaged result of all folds for each dataset for each topological measure. Experiments are performed with the NC LH/RH dataset.}
	\label{fig:numerical_results2} 
\end{figure}

\begin{figure}[h]
	\centering
        {\includegraphics[trim={0 13.8cm 11cm 0cm},clip,width=0.98\textwidth,page=1]{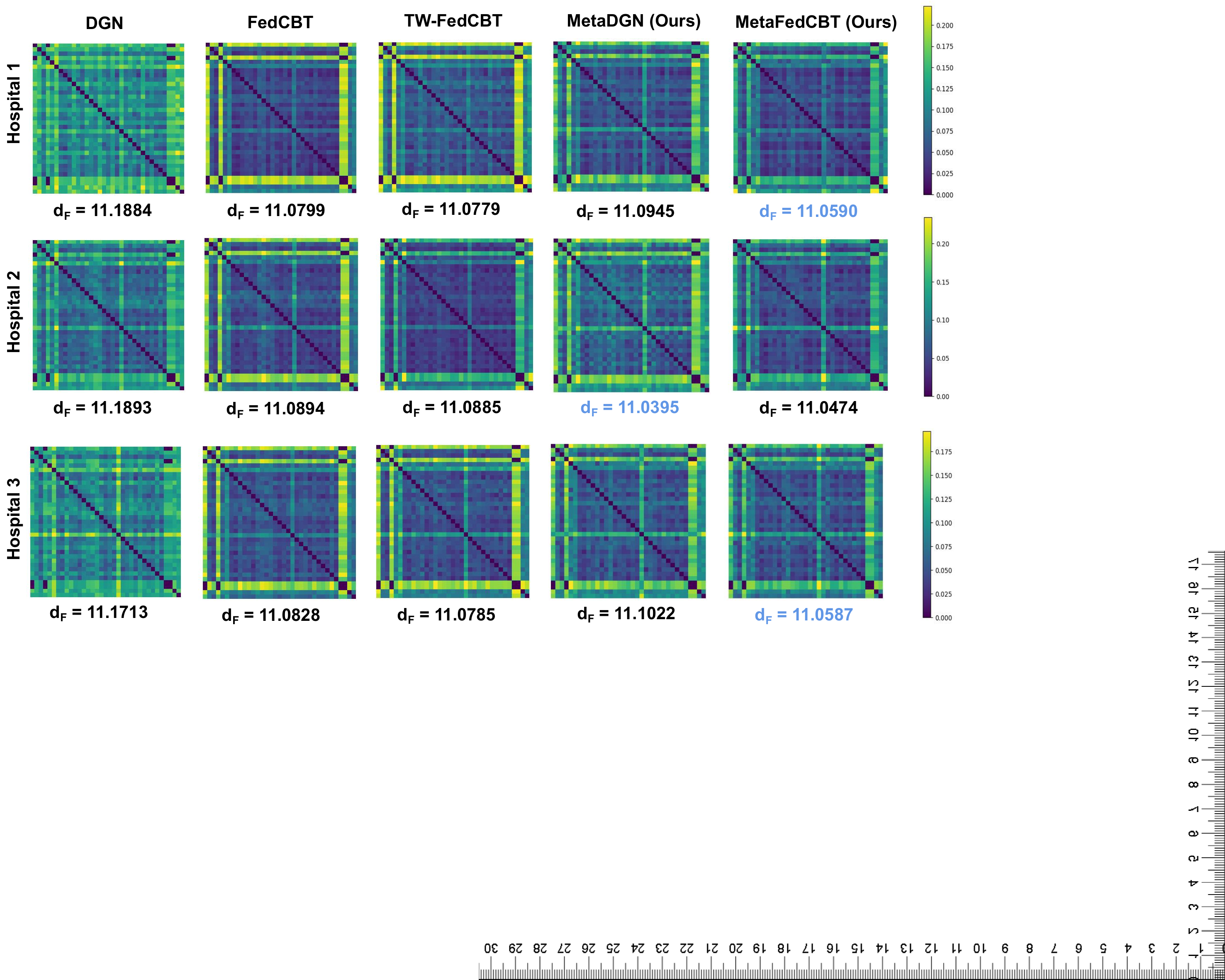}}
        \vspace{-3pt}
 \caption{Visual comparison of CBTs given by \ourmodel{}, the comparison models, and the ablated versions. Results are based on fold one of the ASD LH dataset across three hospitals.}
	\label{fig:visual1} 
\end{figure}

\begin{figure}[h]
	\centering
        {\includegraphics[trim={0 14cm 17cm 0cm},clip,width=0.99\textwidth,page=2]{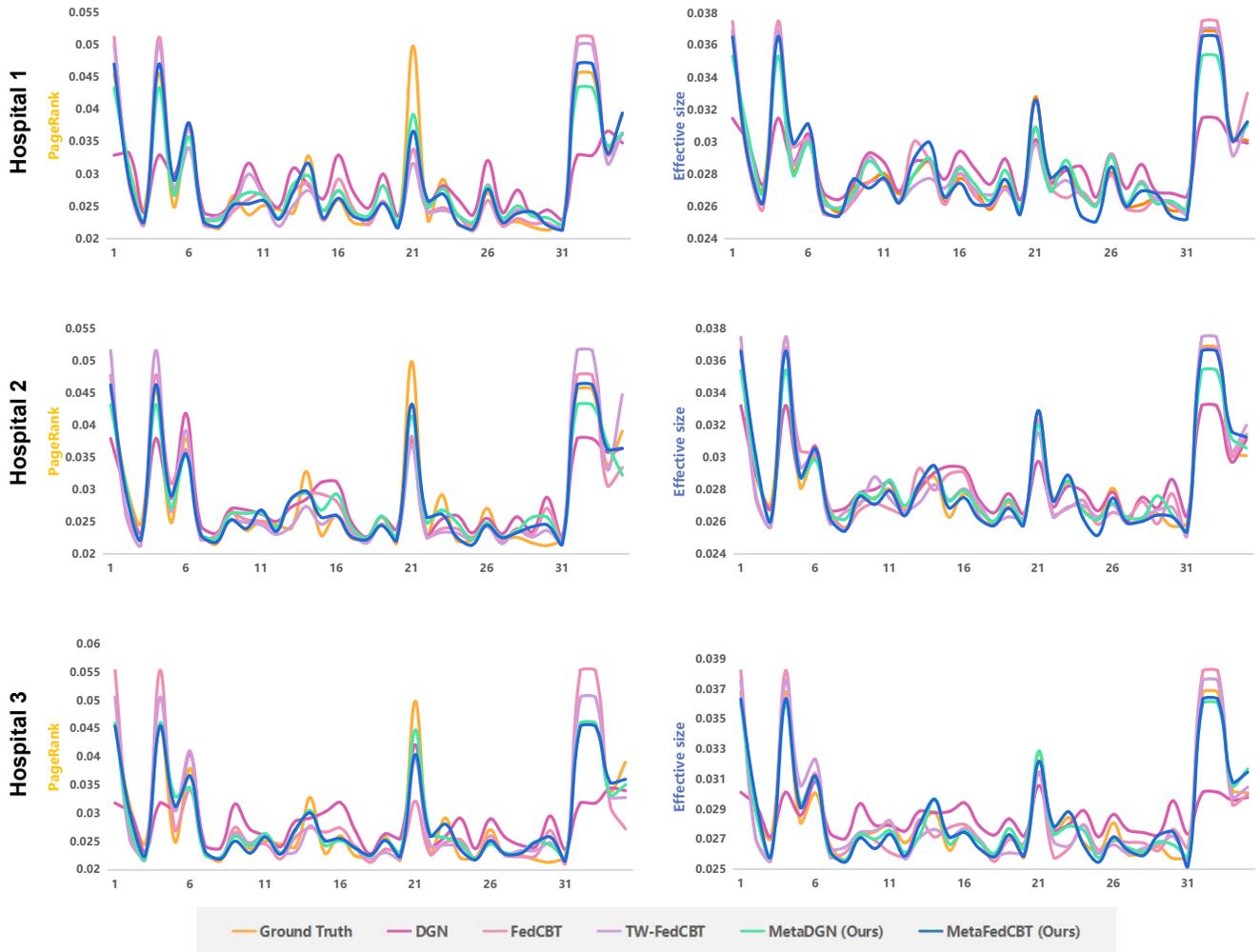}} 
        \vspace{-6pt}
 \caption{Visual comparison of topological distributions of CBTs given by \ourmodel{}, the comparison models, and the ablated versions, measured by PageRank and effective size. Results are based on fold one of the ASD LH dataset across three hospitals.}
	\label{fig:visual2} 
\end{figure}

\noindent Due to the limit of page length, not all results are included in the main paper.
We report additional experimental results in the supplementary materials.
As shown in Figs.~\ref{fig:numerical_results}, \ref{fig:numerical_results2}, \ref{fig:visual1}, \ref{fig:visual2}, and Table~\ref{tab:classification_results}, our \ourmodel{} consistently outperforms existing methods and achieves the best performance.

\vspace{1cm}
\begin{figure}[h]
	\centering
    \includegraphics[trim={0cm 3.5cm 13cm 0cm},clip,width=0.99\textwidth,page=1]{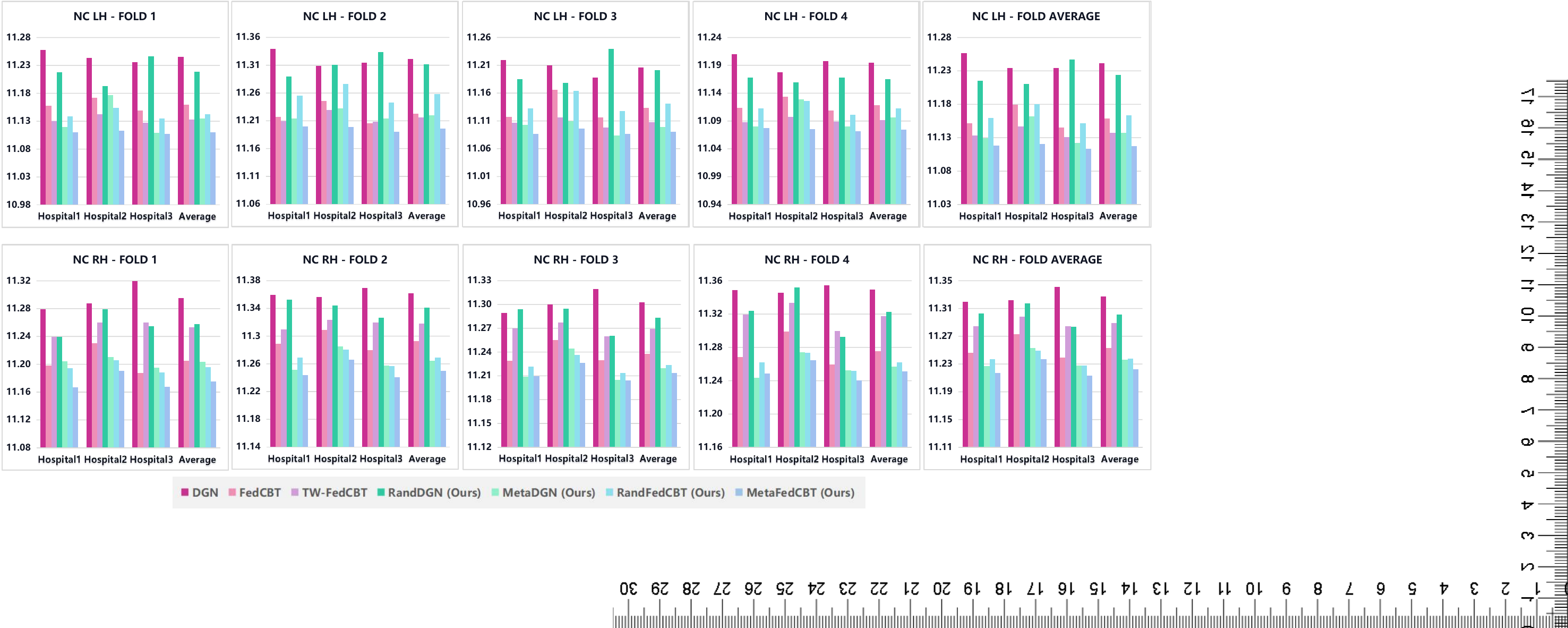}
    \vspace{-6pt}
	\caption{Quantitative comparison of the centeredness of CBTs given by \ourmodel{}, the comparison models, and the ablated versions. The Frobenius distance is adopted for evaluation. Experiments are performed with the NC LH/RH dataset. We also report the $p$-values using a two-tailed paired t-test between \ourmodel{} and each comparison method. If the $p$-value is less than 0.05, it is flagged with one star (*). If the p-value is less than 0.01, it is flagged with 2 stars (**). If the p-value is less than 0.001, it is flagged with 3 stars (***). Lastly, if the p-value is less than 0.0001, it is flagged with (****).}
	\label{fig:numerical_results} 
\end{figure}

\begin{table*}[ht]
	\renewcommand{\arraystretch}{1.3}
	\centering
	\caption{{Testing classification results of SVM classifiers trained with a single CBT from each class of each fold of each hospital. We report the accuracy, precision, recall, and F1 score of the NC-ASD RH classification task over the four cross-validation folds across three hospitals.}}
	\resizebox{\linewidth}{!}{
		\belowrulesep=0pt
		\aboverulesep=0pt
		\begin{tabular}{c|c|cccc|cccc|cccc|cccc}
			\toprule[2pt]
			\multicolumn{1}{c}{} & \multicolumn{1}{c|}{} & \multicolumn{4}{c|}{\textbf{DGN}} & \multicolumn{4}{c|}{\textbf{FedCBT}} & \multicolumn{4}{c|}{\textbf{TW-FedCBT}} & \multicolumn{4}{c}{\textbf{MetaFedCBT (Ours)}} \\
			\cmidrule{3-18}    
			\multicolumn{1}{c}{} & \multicolumn{1}{c|}{} & \multicolumn{1}{c}{Acc} & \multicolumn{1}{c}{Prec} & \multicolumn{1}{c}{Rec} & \multicolumn{1}{c|}{F1} & \multicolumn{1}{c}{Acc} & \multicolumn{1}{c}{Prec} & \multicolumn{1}{c}{Rec} & \multicolumn{1}{c|}{F1} & \multicolumn{1}{c}{Acc} & \multicolumn{1}{c}{Prec} & \multicolumn{1}{c}{Rec} & \multicolumn{1}{c|}{F1} & \multicolumn{1}{c}{Acc} & \multicolumn{1}{c}{Prec} & \multicolumn{1}{c}{Rec} & \multicolumn{1}{c}{F1} \\
			\midrule[1.5pt]

			\multicolumn{1}{c|}{\multirow{5}[3]{*}{Hospital 1}} & FOLD 1 & 0.658  & 0.667  & 0.438  & 0.528  & 0.945  & 0.967  & 0.906  & 0.935  & 0.493  & 0.400  & 0.313  & 0.351  & 0.932  & 0.935  & 0.906  & 0.921  \\
			      & FOLD 2 & 0.606  & 0.571  & 0.387  & 0.462  & 0.958  & 1.000  & 0.903  & 0.949  & 0.620  & 0.577  & 0.484  & 0.526  & 0.859  & 0.800  & 0.903  & 0.848  \\
			     & FOLD 3 & 0.652  & 0.568  & 0.724  & 0.636  & 0.913  & 0.897  & 0.897  & 0.897  & 0.667  & 0.615  & 0.552  & 0.582  & 0.986  & 1.000  & 0.966  & 0.982  \\
			      & FOLD 4 & 0.618  & 0.532  & 0.862  & 0.658  & 0.824  & 0.758  & 0.862  & 0.806  & 0.618  & 0.533  & 0.828  & 0.649  & 0.882  & 0.800  & 0.966  & 0.875  \\
			\cmidrule{2-18}               & Avg.  & 0.633  & 0.584  & 0.603  & 0.571  & 0.910  & $\pmb{0.905}$ & 0.892  & 0.897  & 0.599  & 0.531  & 0.544  & 0.527  & $\pmb{0.915}$ & 0.884  & $\pmb{0.935}$ & $\pmb{0.907}$ \\
			\cmidrule{1-18}          \multicolumn{1}{c|}{\multirow{5}[4]{*}{Hospital 2}} & FOLD 1 & 0.534  & 0.476  & 0.541  & 0.533  & 0.740  & 0.710  & 0.688  & 0.698  & 0.425  & 0.420  & 0.625  & 0.512  & 0.740  & 0.945  & 0.906  & 0.935  \\
		      & FOLD 2 & 0.606  & 0.538  & 0.677  & 0.600  & 0.648  & 0.615  & 0.516  & 0.561  & 0.676  & 0.682  & 0.484  & 0.566  & 0.845  & 0.885  & 0.742  & 0.807  \\
			     & FOLD 3 & 0.768  & 0.667  & 0.897  & 0.765  & 0.696  & 0.700  & 0.483  & 0.571  & 0.478  & 0.426  & 0.690  & 0.526  & 0.855  & 0.828  & 0.828  & 0.828  \\
			    & FOLD 4 & 0.706  & 0.629  & 0.759  & 0.688  & 0.647  & 0.609  & 0.483  & 0.538  & 0.544  & 0.464  & 0.448  & 0.456  & 0.779  & 0.769  & 0.690  & 0.727  \\
			\cmidrule{2-18}                 & Avg.  & 0.653  & 0.577  & 0.718  & 0.646  & 0.683  & 0.658  & 0.542  & 0.592  & 0.531  & 0.498  & 0.562  & 0.515  & $\pmb{0.805}$ & $\pmb{0.857}$& $\pmb{0.791}$& $\pmb{0.824}$ \\
			\cmidrule{1-18}          \multicolumn{1}{c|}{\multirow{5}[4]{*}{Hospital 3}} & FOLD 1 & 0.534  & 0.481  & 0.813  & 0.605  & 0.699  & 0.632  & 0.750  & 0.686  & 0.507  & 0.467  & 0.875  & 0.609  & 0.808  & 0.737  & 0.875  & 0.800  \\
			      & FOLD 2 & 0.648  & 0.714  & 0.323  & 0.444  & 0.620  & 0.556  & 0.645  & 0.597  & 0.465  & 0.419  & 0.581  & 0.466  & 0.761  & 0.733  & 0.710  & 0.721  \\
		     & FOLD 3 & 0.638  & 0.583  & 0.483  & 0.528  & 0.783  & 0.733  & 0.759  & 0.746  & 0.522  & 0.464  & 0.897  & 0.612  & 0.884  & 0.889  & 0.828  & 0.857  \\
			      & FOLD 4 & 0.603  & 0.571  & 0.276  & 0.372  & 0.735  & 0.657  & 0.793  & 0.719  & 0.500  & 0.449  & 0.759  & 0.564  & 0.824  & 0.774  & 0.828  & 0.800  \\
			\cmidrule{2-18}             & Avg.  & 0.606  & 0.588  & 0.473  & 0.487  & 0.709  & 0.644  & 0.737  & 0.687  & 0.498  & 0.450  & 0.778  & 0.563  & $\pmb{0.819}$ & $\pmb{0.783}$ & $\pmb{0.810}$ & $\pmb{0.795}$ \\
			\bottomrule[2pt]
		\end{tabular}%
		\label{tab:classification_results}}
\end{table*}%